\documentclass[lettersize,journal]{IEEEtran}
\usepackage{amsmath,amsfonts}
\usepackage{algorithmic}
\usepackage{algorithm}
\usepackage{array}
\usepackage[caption=false,font=normalsize,labelfont=sf,textfont=sf]{subfig}
\usepackage{textcomp}
\usepackage{stfloats}
\usepackage{url}
\usepackage{verbatim}
\usepackage{graphicx}
\usepackage{cite}
\usepackage[table,xcdraw]{xcolor}
\usepackage{multirow}
\usepackage{makecell}
\usepackage{booktabs}
\newcommand{\bestnum}[1]{\cellcolor{best}\bfseries #1}
\newcommand{\second}[1]{\cellcolor{hicell}#1}

\definecolor{cvprblue}{rgb}{0.21,0.49,0.74}

\usepackage{color}
\usepackage{bm}
\usepackage{flushend}
\usepackage{amssymb}
\usepackage{listings}
\usepackage{colortbl}
\usepackage{subcaption}
\usepackage{pifont}
\newcommand{\cmark}{\ding{51}}

\definecolor{hirow}{HTML}{F8D7DA}    % light pink (row highlight)
\definecolor{hicell}{HTML}{FFF3CD}   % light yellow (cell highlight)
\definecolor{best}{HTML}{FDEDEE}     % stronger pink for "ours"

\definecolor{mygray}{gray}{.9}

\usepackage{arydshln}

\begin{document}

\title{Lite Any Stereo V2: Faster and Stronger \\ Efficient Zero-Shot Stereo Matching}

\author{{Junpeng Jing}, {Ronglai Zuo}, {Zhelun Shen}, {Shangchen Zhou}, {Rolandos Alexandros Potamias}, \\
{Stefanos Zafeiriou}, {Krystian Mikolajczyk}, {Jiankang Deng}
\thanks{J. Jing, R. Zuo, Z. Shen, S. Zhou, R. A. Potamias, S. Zafeiriou, K. Mikolajczyk, and J. Deng are with Imperial College London, London SW7 2AZ, United Kingdom. E-mail: \{j.jing23; r.zuo; zhelun.shen25; s.zhou1; r.potamias; k.mikolajczyk; s.zafeiriou; j.deng16\}@imperial.ac.uk. }
\thanks{Corresponding author: Jiankang Deng}
}

\maketitle

\begin{abstract}
Recent advances in stereo matching have achieved remarkable accuracy, but often rely on large models, heavy computation, or additional foundation priors, making them difficult to deploy on resource-constrained platforms. In contrast, efficient stereo models offer faster inference but are commonly considered less capable of strong zero-shot generalization. In this paper, we challenge this assumption by introducing \textbf{Lite Any Stereo V2 (LAS2)}, an ultra-fast model series designed for efficient zero-shot stereo matching. LAS2 is developed from both architecture and training perspectives. Architecturally, we revisit efficient stereo design under practical deployment settings and propose a 2D-only cost aggregation framework, optimized for real inference latency rather than theoretical MACs alone. For training, we develop a three-stage strategy that combines synthetic supervision, self-distillation, and real-world knowledge distillation. To improve the reliability of real-world pseudo supervision, we further introduce pseudo-label filtering and an error-clamping operation, enabling smoother synthetic-to-real transfer. We instantiate LAS2 as a family of models, including feed-forward variants for different efficiency budgets and an iterative variant for higher accuracy. Extensive experiments show that LAS2 achieves state-of-the-art accuracy among efficient stereo methods while maintaining significantly lower latency. Specifically, LAS2-M consistently outperforms our previous SOTA feed-forward efficient method LAS across four real-world benchmarks, while running 1.6$\times$ faster on H200 and 1.9$\times$ faster on Orin 8G. LAS2-H further achieves stronger overall zero-shot performance than the iterative method Fast-FoundationStereo, with 1.8$\times$ and 2.7$\times$ faster inference on H200 and Orin, respectively. The project page, demos, and code are available at \textcolor{magenta}{https://tomtomtommi.github.io/LiteAnyStereoV2/}.
\end{abstract}

\begin{IEEEkeywords}
Stereo matching, efficient stereo matching, dense correspondence, zero-shot generalization.
\end{IEEEkeywords}

\section{Introduction}
\label{sec:intro}
\IEEEPARstart{F}{rom} the foundational work of \cite{marr1988cooperative} to classical advances such as \cite{taniai2017continuous}, stereo vision has progressed for decades through a wide range of algorithmic developments. In the past decade, deep learning has brought a substantial leap in accuracy, creating the impression that stereo matching is approaching maturity on standard benchmarks. However, this progress has often been driven by increasingly large and computationally expensive models, making high-performance stereo difficult to deploy on resource-constrained platforms.

\begin{figure}[t]
   \begin{center}
   \includegraphics[width=1\linewidth]{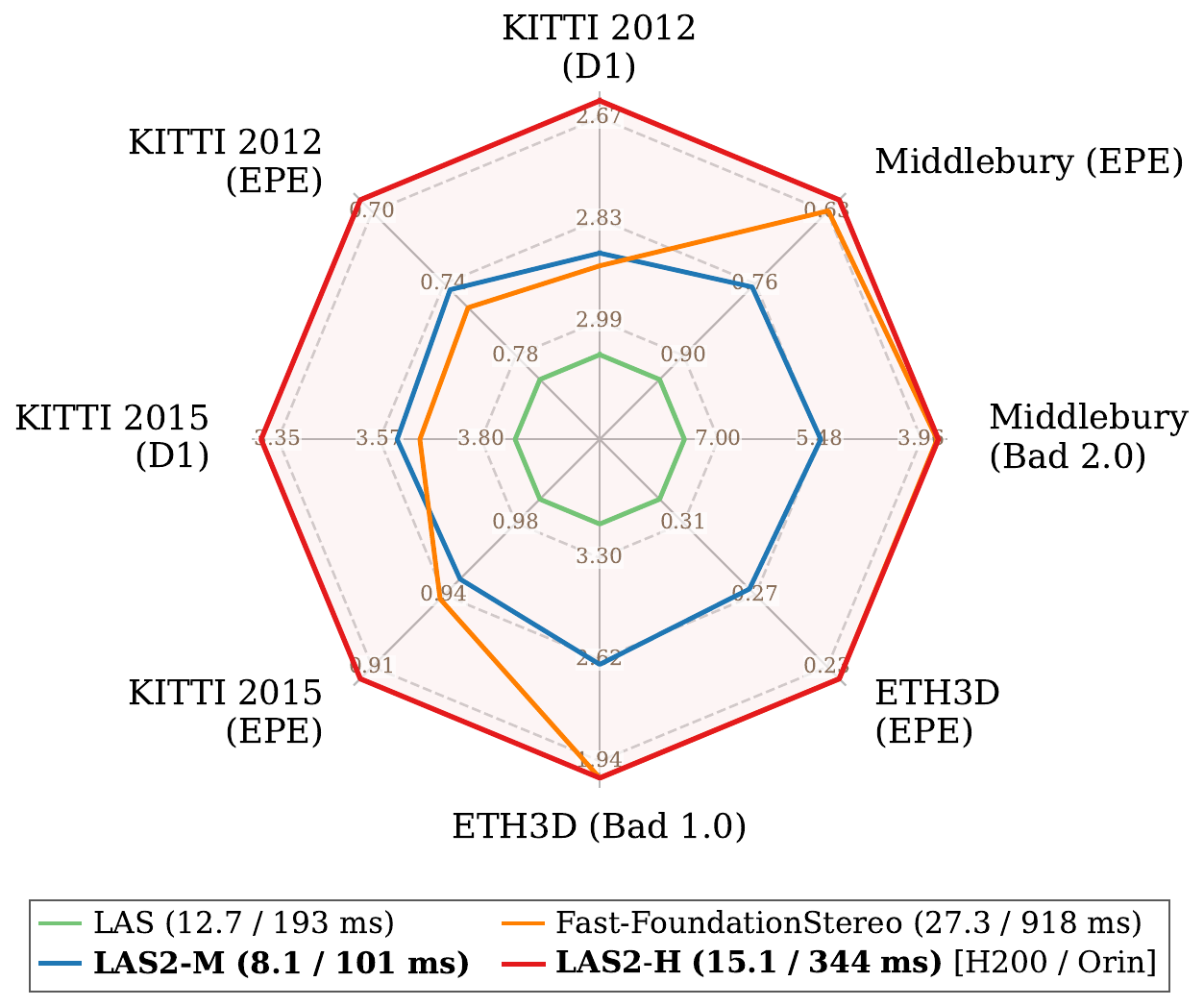}
   \end{center}
   \caption{Zero-shot performance on four real-world benchmarks. The proposed LAS2-M improves both accuracy and latency over the previous efficient feed-forward SOTA method LAS~\cite{jing2025liteanystereo}, while LAS2-H achieves stronger overall results than the iterative method Fast-FoundationStereo~\cite{wen2026fastfoundationstereo} with substantially faster inference speed. The numbers in parentheses denote latency on H200 and Orin NX 8G, respectively.}
   \label{fig:figure1}
\end{figure}

Learning-based stereo methods \cite{lipson2021raft, li2022practical, xu2023accurate, wang2024selective} have achieved remarkable accuracy and continuously improved results on standard benchmarks \cite{middlebury, eth3d, kitti12, kitti15}. These methods generally prioritize accuracy, often at the cost of substantial computation. More recently, the emergence of foundation models trained on internet-scale data, such as the DepthAnything series \cite{yang2024depth, yang2024depth2}, has further advanced the field. By incorporating monocular depth priors, recent stereo systems \cite{wen2025foundationstereozeroshotstereomatching, cheng2025monstermarrymonodepthstereo, jiang2025defomstereodepthfoundationmodel} have demonstrated strong zero-shot generalization, where a single set of model weights can perform well across diverse scenarios. Despite their impressive performance, these approaches remain primarily accuracy-driven and often require heavy backbones or additional prior networks, limiting their applicability in real-world deployment scenarios.

In contrast, efficiency-oriented approaches~\cite{shamsafar2022mobilestereonet,guo2024lightstereochannelboostneed, xu2025banet} trade accuracy for faster inference and lower resource use, however, the accuracy gap to large stereo models remains significant. This gap may create the impression that lightweight stereo networks inherently lack sufficient capacity for zero-shot generalization. As a result, many efficient models still rely on domain-specific fine-tuning and therefore fall short of being practical off-the-shelf stereo solutions. Some recent methods~\cite{xu2025igev++,cheng2025monsterunifiedstereomatching,wen2026fastfoundationstereo} attempt to improve efficiency by compressing iterative stereo pipelines, but their speed advantages are still mainly observed on high-end GPUs, leaving efficient deployment on edge devices less explored. Other efforts further exploit monocular depth models to synthesize stereo pairs from single real-world images~\cite{guo2024stereo}. Since the synthesized supervision is not derived from real left-right correspondence, it often lacks accurate stereo geometry and reliable fine-grained details. Such data alone remains insufficient for closing the gap between lightweight and accuracy-oriented models.

\begin{figure*}[t]
   \begin{center}
   \includegraphics[width=1\linewidth]{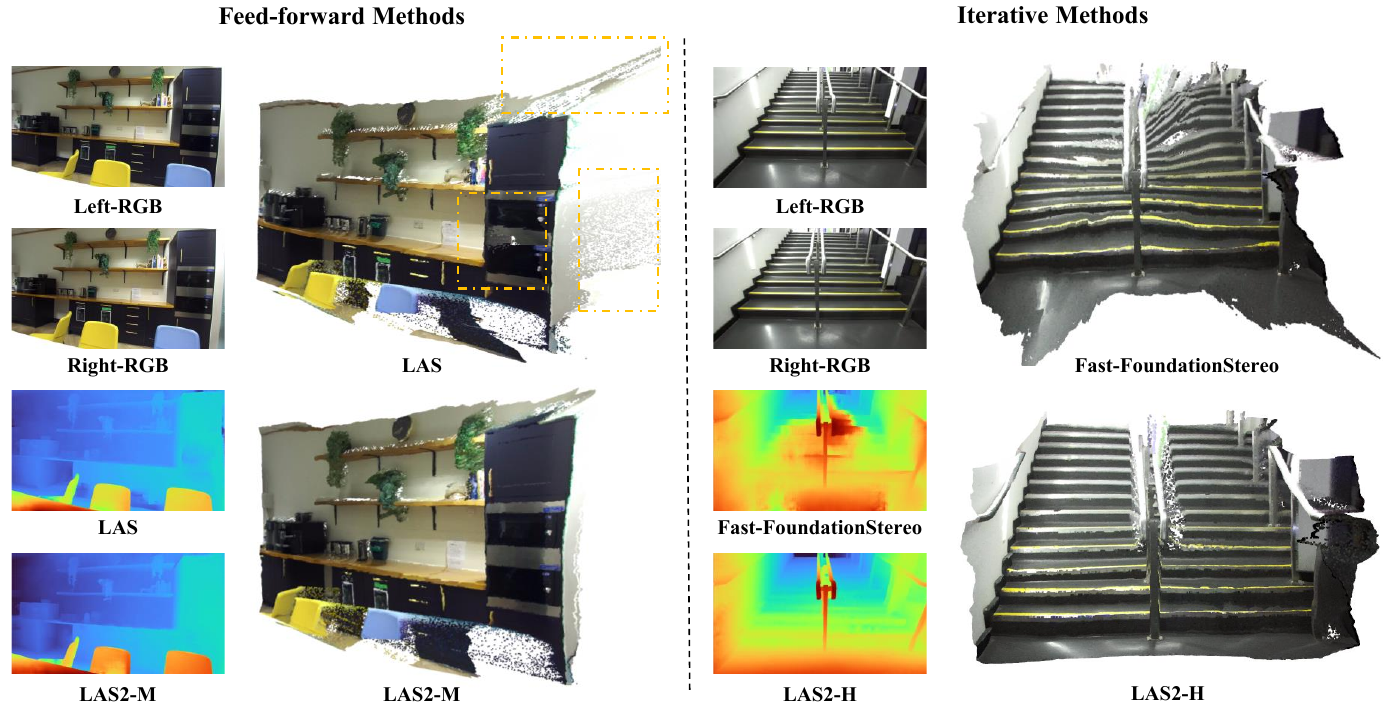}
   \end{center}
    \caption{Zero-shot prediction on in-the-wild stereo images. We compare disparity maps and reconstructed raw metric point clouds without denoising. Compared with LAS~\cite{jing2025liteanystereo} and Fast-FoundationStereo~\cite{wen2026fastfoundationstereo}, the proposed LAS2 produces cleaner disparity estimates and more complete reconstructions, demonstrating strong zero-shot ability while maintaining high efficiency.
    }
   \label{fig:figure2}
\end{figure*}

In this paper, we propose Lite Any Stereo V2 (LAS2), an ultra-fast stereo matching model series designed for zero-shot generalization. LAS2 is developed from two complementary aspects: architecture design and training strategy. For architecture, we introduce a 2D-only convolution-based cost aggregation framework, avoiding the heavy 3D aggregation commonly used in accurate stereo models. We extensively ablate key design choices with a focus on practical latency on GPUs and edge devices, rather than relying only on Multiply-Accumulate Operations (MACs), which often fail to reflect real inference speed. For training, we scale stereo learning to the million-sample level with a carefully designed three-stage strategy. After supervised training on synthetic labeled data, we perform self-distillation to improve robustness to input perturbations. We then exploit real-world unlabeled stereo images, which have so far been underused in stereo matching, through knowledge distillation from strong teacher models. To make real-world pseudo supervision more reliable, we further introduce a pseudo-label filtering mechanism and an error-clamping operation, which help suppress noisy labels and enable smoother synthetic-to-real transfer.

To meet different deployment requirements on GPUs and edge devices, we instantiate LAS2 as a family of models, including S, M, L, and H. The first three variants are feed-forward models with increasing capacity, while LAS2-H adopts an iterative-based framework for higher accuracy. Compared with representative feed-forward and iterative SOTA methods, LAS2 achieves stronger zero-shot performance with significantly lower latency. As shown in Fig.~\ref{fig:figure1}, LAS2-M achieves consistently lower errors across the four real-world benchmarks than the feed-forward method LAS~\cite{jing2025liteanystereo}, while running 1.6$\times$ and 1.9$\times$ faster on H200 and Orin, respectively. LAS2-H further pushes the accuracy frontier: compared with the iterative method Fast-FoundationStereo~\cite{wen2026fastfoundationstereo}, it achieves stronger overall performance while running 1.8$\times$ and 2.7$\times$ faster on H200 and Orin, respectively. These results show that lightweight stereo models can achieve strong zero-shot generalization without sacrificing deployment efficiency. As shown in Fig.~\ref{fig:figure2}, LAS2 generalizes well to in-the-wild stereo images, producing accurate disparity maps with high efficiency.

Our main contributions are summarized as follows:
\begin{itemize}
\item We present LAS2, an efficient stereo matching model series for zero-shot generalization. LAS2 achieves SOTA accuracy among efficient stereo methods while running significantly faster than representative baselines. It also narrows the gap to accuracy-oriented methods, approaching their performance with much lower latency.

\item We systematically study efficient stereo architecture design under practical deployment settings and develop a purely 2D cost aggregation framework, achieving a favorable accuracy-latency trade-off.

\item We propose a three-stage training strategy that combines synthetic supervision, self-distillation, and real-world knowledge distillation, enabling efficient stereo models to generalize across diverse real-world scenarios.

\item We introduce a pseudo-label filtering mechanism to improve the reliability of real-world pseudo supervision, together with an error-clamping operation to facilitate smoother synthetic-to-real transfer.
\end{itemize}

\textit{Differences to conference version:} This work substantially extends our  conference paper~\cite{jing2025liteanystereo}, with the key differences summarized as follows:

\noindent \textbf{Architecture.} We redesign the architecture from the perspective of practical deployment. While the previous version achieved a favorable MAC budget, we find that MACs alone do not reliably reflect real inference speed on modern GPUs and edge devices. We therefore revisit the backbone and aggregation design, moving from the previous hybrid aggregation module to a deployment-friendly 2D-only architecture that substantially improves practical latency.

\noindent \textbf{Training strategy.} We further enhance the three-stage training strategy with pseudo-label filtering and error clamping. These improve the reliability of pseudo supervision, and reduce the negative impact of remaining noisy labels, enabling smoother synthetic-to-real transfer and stronger zero-shot generalization.

\noindent \textbf{Experiments.} Compared with the previous version, LAS2-M reduces the error by 13.7\% across the four real-world benchmarks, while running 1.9$\times$ faster. We also provide a more comprehensive evaluation, including zero-shot and in-domain settings, qualitative comparisons, and latency measurements on different GPUs and edge-device power modes.

\noindent \textbf{Model family.} We expand the original single-model design into a complete LAS2 model family, including feed-forward and iterative variants for different efficiency requirements.

The remainder of this paper is organized as follows. Section \ref{sec: Related Work} reviews existing stereo matching methods. Section \ref{sec: Method} presents the proposed network architecture and training strategy in detail. Section \ref{sec: Experiments} reports the experimental results and comparisons with state-of-the-art methods. Section \ref{sec: Limitations} discusses the limitations and remaining open challenges. Finally, Section \ref{sec: Conclusion} concludes the paper.

\section{Related Work} \label{sec: Related Work}
In this section, we first summarize the development of deep stereo networks, then discuss methods for improving zero-shot generalization, and finally review efficient approaches.

%-------------------------------------------------------------------------
\subsection{Deep Stereo Methods} 
Traditional stereo matching pipelines are usually built upon hand-crafted matching costs, aggregation, disparity selection, and post-processing steps~\cite{kanade1994stereo, hsieh1992performance, scharstein1998stereo}. With the emergence of deep learning, end-to-end stereo networks have gradually become the dominant paradigm since DispNet~\cite{mayer2016large}. Most methods construct a cost volume to encode correspondences between the left and right images. Depending on the formulation, this representation can be a 3D cost volume over height, width, and disparity, or a 4D cost volume that further preserves the feature dimension. The resulting cost volume is then regularized by neural networks to infer disparities.

Existing deep stereo networks can be broadly grouped into feed-forward and iterative methods. Feed-forward approaches~\cite{chang2018pyramid, guo2019group, xu2020aanet, tankovich2021hitnet, guo2023openstereo} usually estimate disparity in a single forward pass by applying cost aggregation and disparity regression. In contrast, iterative methods~\cite{lipson2021raft, li2022practical, Jing_2023_ICCV, xu2023iterative, wang2024selective} repeatedly update disparity predictions through local cost volume lookup and recurrent refinement, leading to better accuracy. Another line of work adopts transformer architectures~\cite{guo2022context, su2022chitransformer, weinzaepfel2023croco, xu2023unifying}, where attention mechanisms are used to model long-range dependencies and global context. Beyond image-based stereo, recent studies also investigate video stereo matching~\cite{karaev2023dynamicstereo, jing2024matchstereovideos, jing2024matchstereovideosbidirectional, jing2025stereovideotemporallyconsistent}, with a particular focus on improving temporal consistency. Despite the strong performance on standard benchmarks~\cite{kitti12, kitti15, middlebury, eth3d}, these models remain sensitive to domain shifts, and their zero-shot generalization to unseen scenes is still limited.

\subsection{Zero-Shot Stereo Methods}
To improve cross-domain robustness, early zero-shot stereo methods mainly focus on learning domain-invariant representations. Representative techniques include domain normalization and non-local graph filtering in DSMNet~\cite{zhang2020domain}, shortcut learning in ITSA~\cite{chuah2022itsa}, stereo contrastive feature learning~\cite{zhang2022revisiting}, hierarchical visual transformation~\cite{chang2023domain}, and masked image modeling~\cite{rao2023masked}. Other methods~\cite{shen2021cfnet, Jing_2023_ICCV, guo2024stereo} also improve robustness under domain shifts through architectural or training-design choices. Monocular foundation models have recently opened a new direction for zero-shot stereo matching. Specifically, models from the Depth Anything series~\cite{yang2024depth, yang2024depth2} provide strong monocular depth priors that can be incorporated into stereo pipelines. By exploiting such priors, several methods~\cite{bartolomei2024stereo, zhang2024learning, zhou2024all, cheng2025monstermarrymonodepthstereo, jiang2025defomstereodepthfoundationmodel, wen2025foundationstereozeroshotstereomatching, jing2025stereovideotemporallyconsistent} have achieved substantially improved zero-shot performance. However, the additional prior module and complex pipelines often introduce considerable computational overhead. Therefore, how to obtain strong zero-shot performance while preserving real-time efficiency remains an important and challenging problem.

\subsection{Efficient Stereo Matching}
Real-time processing is essential for practical applications such as robotics, autonomous driving, and embedded perception. Early efficient methods~\cite{khamis2018stereonet, duggal2019deeppruner, wang2020fadnet} reduce computation by predicting disparities at lower resolutions, but this often sacrifices fine-grained accuracy. Later works seek a better balance between efficiency and accuracy while still relying on 3D cost aggregation. For example, CoEx~\cite{bangunharcana2021correlate} introduces guided cost-volume excitation, BGNet~\cite{xu2021bilateral} improves boundary quality with edge-aware upsampling, and Fast-ACVNet~\cite{xu2022acvnet, xu2023accurate} uses sparse attention to avoid unnecessary high-resolution matching. Another line of research reduces the reliance on expensive 3D convolutions by developing 2D-based alternatives. AANet~\cite{xu2020aanet} performs adaptive cost aggregation with deformable 2D convolutions, HITNet~\cite{tankovich2021hitnet} proposes iterative warping to avoid constructing an explicit dense cost volume, and MobileStereoNet~\cite{shamsafar2022mobilestereonet} adopts lightweight MobileNet-style blocks. Recent methods further improve efficiency through more specialized designs, such as channel-wise enhancement in LightStereo~\cite{guo2024lightstereochannelboostneed} and frequency guided bilateral aggregation in BANet~\cite{xu2025banet}. Although these models are computationally efficient, they are often optimized for specific domains, especially the KITTI benchmarks~\cite{kitti12, kitti15}, and their generalization to diverse unseen scenarios remains limited.

More recently, some methods also explore real-time variants of high-accuracy models. Lite-CREStereo++~\cite{Jing_2023_ICCV} and RT-IGEV++~\cite{xu2025igev++} reduce latency by decreasing channel dimensions and iteration numbers while retaining the core modules of their original models. RT-MonSter++~\cite{cheng2025monsterunifiedstereomatching} replaces the heavy DepthAnythingV2-L~\cite{yang2024depth2} prior with a smaller variant to improve inference speed. Fast-FoundationStereo~\cite{wen2026fastfoundationstereo} combines neural architecture search, structured pruning, and knowledge distillation to obtain a more efficient model while maintaining competitive zero-shot performance. However, these methods are primarily designed for high-end GPUs, and their deployment on edge devices remains difficult, which limits their applicability in resource-constrained scenarios.

\begin{figure*}[t]
   \begin{center}
   \includegraphics[width=1\linewidth]{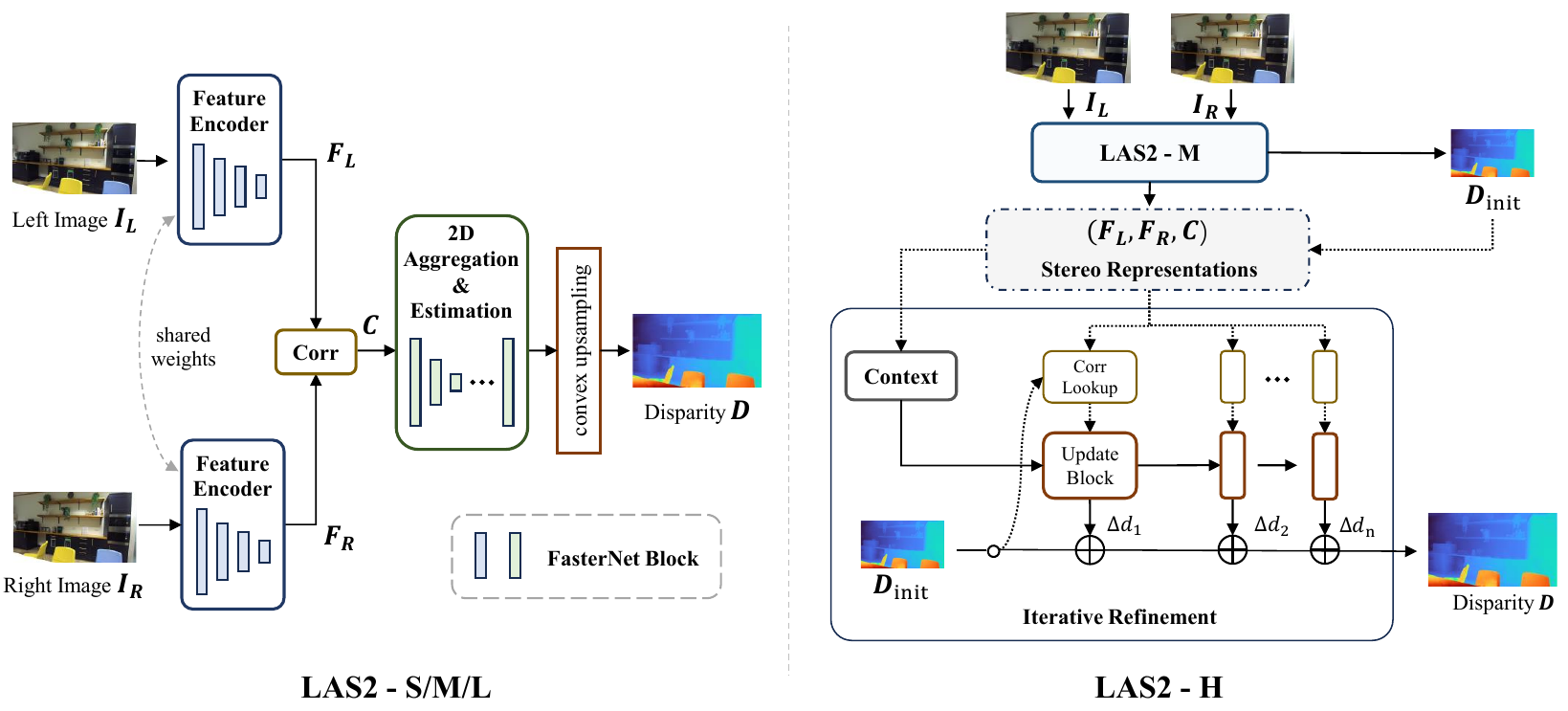}
   \end{center}
   % \vspace{-1.5em}
    \caption{Overview of the proposed Lite Any Stereo V2 (LAS2). \textbf{Left:} LAS2-S/M/L adopt a compact feed-forward pipeline, where shared-weight encoders extract stereo features to construct a correlation volume, followed by 2D cost aggregation and convex upsampling for full-resolution disparity prediction. \textbf{Right:} LAS2-H introduces an iterative refinement pipeline. It uses LAS2-M to provide an initial disparity and intermediate stereo representations, which are further refined by a context-guided recurrent update module to improve accuracy.}
    % \vspace{-.8em}
   \label{fig:framework}
\end{figure*}

\section{Method} \label{sec: Method}

In this section, we introduce the Lite Any Stereo V2 (LAS2) model family. We first present the feed-forward variants, LAS2-S/M/L, in Sec.~\ref{sec: feed_forward framework}. We then describe the iterative high-accuracy variant, LAS2-H, in Sec.~\ref{sec: iterative framework}. Finally, we introduce the proposed training strategy in Sec.~\ref{sec: training strategy}.

\subsection{Feed-forward Framework: LAS2-S/M/L}  \label{sec: feed_forward framework}
As shown in the left part of Fig.~\ref{fig:framework}, the feed-forward LAS2-S/M/L framework consists of four main stages: feature extraction, correlation, cost aggregation, and disparity estimation.

\noindent \textbf{Feature Extraction.} Recent stereo methods~\cite{jiang2025defomstereodepthfoundationmodel, cheng2025monstermarrymonodepthstereo, wen2025foundationstereozeroshotstereomatching} have achieved remarkable performance by leveraging monocular depth features from Depth Anything (DA)~\cite{yang2024depth, yang2024depth2}. Although these depth priors are powerful, even the smallest DA-S variant introduces substantial computational overhead, making it unsuitable for an efficiency-oriented stereo model. We therefore adopt a conventional ImageNet-pretrained backbone for feature extraction, following efficient stereo designs~\cite{guo2024lightstereochannelboostneed,xu2025banet}. While existing lightweight stereo methods, such as LAS \cite{jing2025liteanystereo},  often employ MobileNetV2~\cite{sandler2018mobilenetv2} due to its compact channel configuration and low theoretical complexity, MACs do not always translate to practical latency. We thus use FasterNet~\cite{chen2023run}, which has slightly higher MACs but achieves faster inference in practice, making it better aligned with our deployment-oriented design.

Specifically, given a pair of rectified stereo images $\{\mathbf{I}_{L}, \mathbf{I}_{R}\} \in \mathbb{R}^{H\times W\times 3}$, we use two weight-sharing feature encoders to extract multi-scale feature pyramids $\{\mathbf{F}_{L}^s\}$ and $\{\mathbf{F}_{R}^s\}$, where $s \in \left\{\frac{1}{4}, \frac{1}{8}, \frac{1}{16}, \frac{1}{32}\right\}$ denotes the downsampling ratio. To provide a unified spatial resolution for subsequent matching, features from all scales are upsampled to $\tfrac{1}{4}$ resolution using residual upsampling blocks, following~\cite{guo2024lightstereochannelboostneed}.

\noindent \textbf{Correlation.} Given the left and right feature maps $\mathbf{F}_{L}^{\frac{1}{4}}$ and $\mathbf{F}_{R}^{\frac{1}{4}}$, we construct a cost volume $\mathbf{C}$ over the disparity range $[0, D_{\mathrm{max}} / 4]$ as:
\begin{equation}
    \mathbf{C}(d, h, w) = \frac{1}{N_c} \left\langle \mathbf{F}_{L}^{\frac{1}{4}}(h, w), \ \mathbf{F}_{R}^{\frac{1}{4}}(h, w-d) \right\rangle ,
    \label{eq:1}
\end{equation}
where $D_{\max}$ denotes the predefined maximum disparity value, $\langle \cdot, \cdot \rangle$ denotes the inner product, $N_c$ is the number of channels, and $(h,w)$ represents the pixel location.

\noindent \textbf{Cost Aggregation.} Our previous approach, LAS~\cite{jing2025liteanystereo}, adopts a hybrid 3D-2D aggregation module to combine geometric reasoning with efficient spatial refinement. In this design, the 3D component aggregates information jointly along the disparity and spatial dimensions, while the 2D component further refines the cost representation in the image plane. Although this hybrid design improves geometric modeling, it also introduces additional computational overhead, which is less noticeable on high-end GPUs but becomes significant on edge devices. Motivated by practical deployment efficiency, LAS2 removes the 3D aggregation component and adopts a 2D-only cost aggregation design, following recent efficient stereo methods~\cite{shamsafar2022mobilestereonet,guo2024lightstereochannelboostneed,xu2025banet}.

\begin{figure*}[t]
   \begin{center}
   \includegraphics[width=1\linewidth]{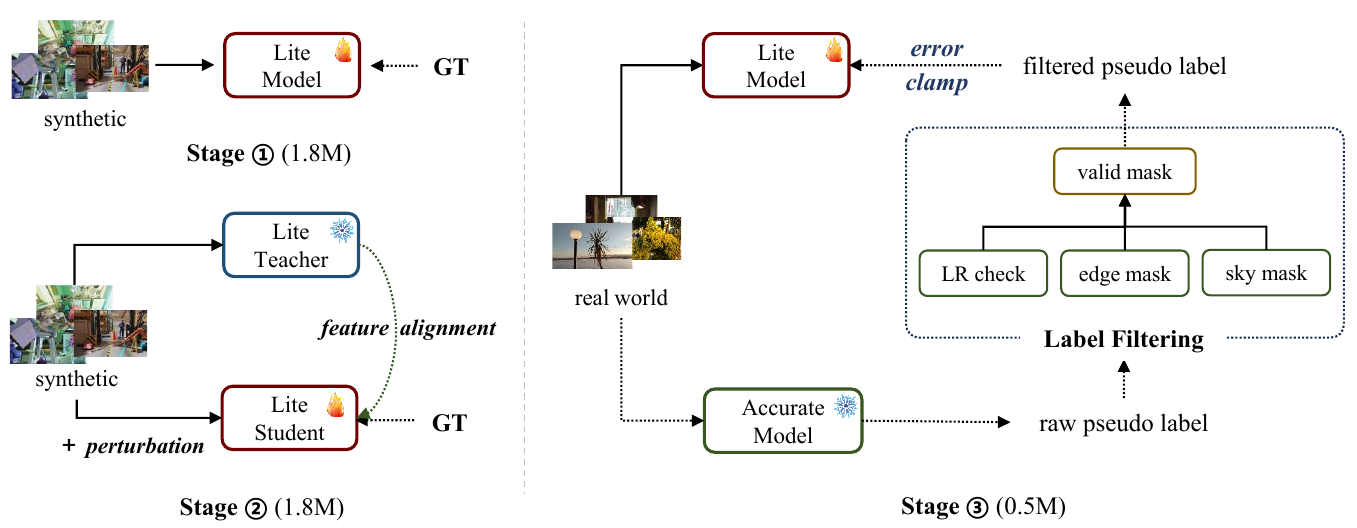}
   \end{center}
   % \vspace{-1.2em}
    \caption{Overview of the proposed three-stage training strategy. \textbf{Stage \ding{172}:} The lite model is trained using a standard supervised setup on a mixture of synthetic datasets including 1.8M labeled stereo image pairs. \textbf{Stage \ding{173}:} Self-distillation is employed, where both teacher and student models are initialized from the Stage \ding{172} weights. The teacher receives clean data, while the student is fed perturbed inputs to encourage learning of domain-invariant representations via feature alignment. \textbf{Stage \ding{174}:}  The lite model is further fine-tuned on 0.5M unlabeled real-world stereo pairs using pseudo labels generated by a frozen accurate model. The raw pseudo labels are refined by label filtering, which combines LR left-right consistency check, edge masks, and sky masks to produce valid masks, and are then used with an error-clamped loss for robust supervision.}
    % \vspace{-.6em}
   \label{fig:training_strategy}
\end{figure*}

Specifically, we adopt a U-Net-style aggregation network following \cite{guo2024lightstereochannelboostneed}. The cost volume $\mathbf{C}$ is progressively aggregated over three resolution levels using strided residual layers, and is then restored to the original cost resolution through two transposed-convolution upsampling layers with skip connections. Consistent with our backbone design, the residual layers are implemented with FasterNet-style blocks. We also retain the attention mechanism from \cite{guo2024lightstereochannelboostneed}. The resulting aggregated cost volume $\mathbf{C}_{agg}$ is then used for disparity estimation.

The three feed-forward variants, S, M, L, use the same feature extraction and correlation design, while scaling the model capacity by varying the depth of the cost aggregation module. For LAS2-S, we use ${1, 2, 4}$ layers at the $\frac{1}{4}$, $\frac{1}{8}$, and $\frac{1}{16}$ resolutions, respectively, in both the encoder and decoder. LAS2-M and LAS2-L use the corresponding encoder-decoder layer configurations of ${4, 8, 16}$ and ${8, 16, 32}$, respectively.

\noindent \textbf{Disparity Estimation.} Similar to other efficient methods \cite{guo2024lightstereochannelboostneed, xu2025banet}, we apply the soft-argmax operation to regress the disparity map $\mathbf{d}$ at $\frac{1}{4}$ scale:
\begin{equation}
    \mathbf{d} = \sum_{d=0}^{D_{\max}/4} d \times \sigma(\mathbf{C}_{\text{agg}}(d)),
    \label{eq:2}
\end{equation}
where $\sigma(\cdot)$ is a softmax layer. Convex upsampling is then used to upsample $\mathbf{d}$ to the full-resolution $\mathbf{D} \in \mathbb{R}^{H \times W}$.

\subsection{Iterative Framework: LAS2-H}
\label{sec: iterative framework}
In Fig.~\ref{fig:framework} right, we further introduce an iterative framework, termed LAS2-H. Unlike most IGEV-style iterative pipelines~\cite{xu2025igev++}, which use 3D convolutions to regularize the cost or geometry volume for initial disparity estimation and subsequent refinement, LAS2-H adopts the 2D cost aggregation module introduced above to produce the initial disparity. This improves efficiency and enables the model to reuse the pretrained LAS2-M weights.

Specifically, given left and right images $\{\mathbf{I}_{L}, \mathbf{I}_{R}\}$, LAS2-M first predicts an initial disparity $\mathbf{d}_{init}$ and produces intermediate stereo representations $\{\mathbf{F}_{L}, \mathbf{F}_{R}, \mathbf{C}_{g}\}$. Although LAS2-M is using the standard correlation volume from Eq.~\ref{eq:1}, in LAS2-H we first construct its group-wise form:
\begin{equation}
\mathbf{C}_g(g,d,h,w)=\frac{1}{N_c/N_g}\left\langle \mathbf{F}_{L}^{g}(h,w),\mathbf{F}_{R}^{g}(h,w-d)\right\rangle,
\label{eq:3}
\end{equation}
where $N_g$ is the number of groups from the number of feature channels $N_c$. The standard correlation volume used by the LAS2-M aggregation module can be directly obtained by averaging $\mathbf{C}_g$ over the group dimension:
\begin{equation}
\mathbf{C}(d,h,w)=\frac{1}{N_g}\sum_{g=1}^{N_g}\mathbf{C}_g(g,d,h,w).
\label{eq:4}
\end{equation}
This transformation is equivalent to Eq.~\ref{eq:1}, and therefore keeps the input format of the LAS2-M unchanged. As a result, LAS2-H can reuse its pretrained weights while retaining the group-wise volume $\mathbf{C}_g$ for iterative geometry lookup.

\begin{figure*}[t]
   \begin{center}
   \includegraphics[width=1\linewidth]{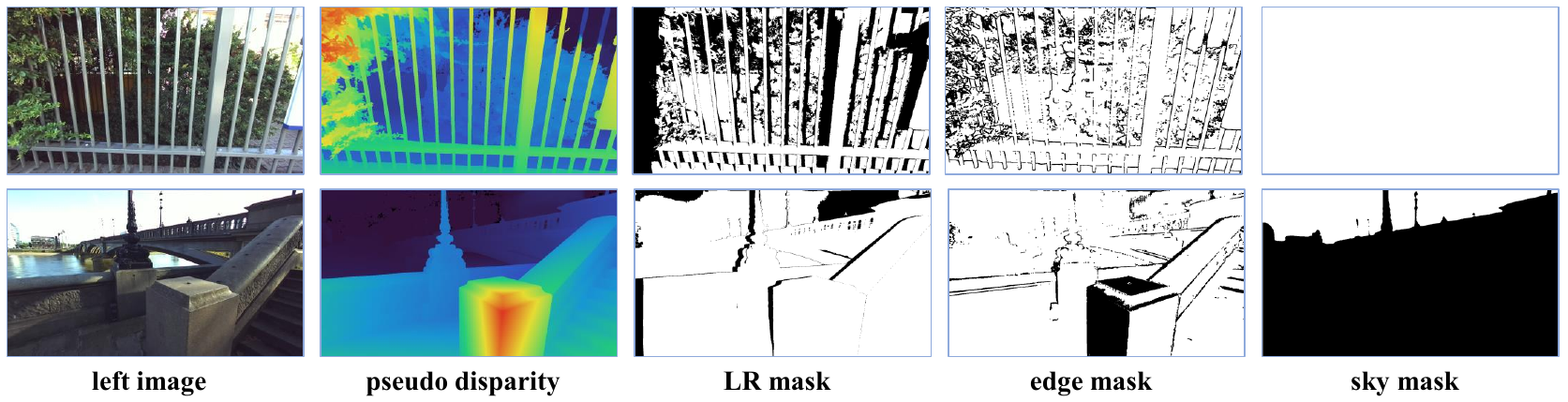}
   \end{center}
   \caption{
    Visualization of validity cues for pseudo-label filtering.
    The left-right consistency mask, edge mask, and sky segmentation mask provide complementary reliability cues for filtering pseudo disparities.
    They remove noisy supervision around occlusions, depth discontinuities, and sky/background regions, leading to more reliable pseudo-label training on real-world stereo data.
    }
   \label{fig:stage3_mask}
\end{figure*}

\begin{table}[t] \addtolength{\tabcolsep}{0pt}
\caption{Overview of the real-world stereo datasets used for model training. These datasets comprise approximately 0.5M stereo image pairs in total and cover diverse indoor and outdoor real-world scenes.}
% \vspace{-.6em}
\centering
\small
\begin{tabular}{lccccr}
\toprule
\textbf{Dataset} & \textbf{Indoor} & \textbf{Outdoor} & \textbf{MPix} & \textbf{Images} \\
\midrule
Flickr1024~\cite{Flickr1024} & \cmark & \cmark  & 0.73 &  1K \\
InStereo2k~\cite{bao2020instereo2k} & \cmark &  & 0.93 & 2K \\
Holopix50K~\cite{hua2020holopix50k} & \cmark & \cmark  & 0.74 &  49K \\
Driving Stereo~\cite{yang2019drivingstereo} &  & \cmark  & 0.40 &  174K \\
SouthKenSV~\cite{jing2024matchstereovideosbidirectional} & \cmark & \cmark & 0.92 &  113K \\
UASOL \cite{bauer2019uasol} &  & \cmark & 2.74 &   156K \\
\bottomrule
\end{tabular}
% \vspace{-.6em}
\label{tab:dataset_overview}
\end{table}

For iterative refinement, we follow a compact recurrent update design. Starting from $\mathbf{d}_{0}=\mathbf{d}_{init}$, the model progressively refines the disparity using a combined geometry volume and image context features. The geometry volume contains the retained group-wise cost volume $\mathbf{C}_g$ and an all-pairs correlation volume $\mathbf{C}_a$ \cite{lipson2021raft}. At each iteration $k=1,\ldots,n$, the model retrieves local geometry features around the current disparity $\mathbf{d}_{k-1}$, fuses them with the context feature $\mathbf{c}$ extracted from the left image, and updates the hidden state $\mathbf{h}_k$ through a selective ConvGRU \cite{wang2024selective}:
\begin{align}
\mathbf{G}_k &= \operatorname{Lookup}(\mathbf{C}_g,\mathbf{C}_a,\mathbf{d}_{k-1}), \label{eq:5} \\
\mathbf{x}_k &= [\mathrm{Encoder}_{g}(\mathbf{G}_k),\mathrm{Encoder}_{d}(\mathbf{d}_{k-1}),\mathbf{d}_{k-1},\mathbf{c}], \label{eq:6} \\
\hat{\mathbf{h}}_{k}^{s} &= \mathrm{ConvGRU}_{1 \times 1}(\mathbf{h}_{k-1},\mathbf{x}_k), \label{eq:7} \\
\hat{\mathbf{h}}_{k}^{l} &= \mathrm{ConvGRU}_{3 \times 3}(\mathbf{h}_{k-1},\mathbf{x}_k), \label{eq:8} \\
\mathbf{h}_{k} &= \mathbf{A}\odot\hat{\mathbf{h}}_{k}^{s}+(1-\mathbf{A})\odot\hat{\mathbf{h}}_{k}^{l}, \label{eq:9} \\
\mathbf{d}_{k} &= \mathbf{d}_{k-1}+\mathrm{Head}_{d}(\mathbf{h}_{k}). \label{eq:10}
\end{align}
Here, $\operatorname{Lookup}(\cdot)$ denotes local correlation lookup \cite{lipson2021raft}, which samples both $\mathbf{C}_g$ and $\mathbf{C}_a$ over multiple pyramid levels. $\mathbf{A}$ is the spatial attention map~\cite{wang2024selective}. The disparity estimate $\mathbf{d}_k$ at each iteration is upsampled to full resolution $\mathbf{D}_{k}$ via convex upsampling, and the final prediction is given by $\mathbf{D}_{n}$. In practice, the total number of iterations is set to $n=4$. Both the hidden state and context feature have 64 channels. $\mathrm{Encoder}_{g}$ and $\mathrm{Encoder}_{d}$ each consist of two convolutional layers.

\begin{figure}[t]
   \begin{center}
   \includegraphics[width=1\linewidth]{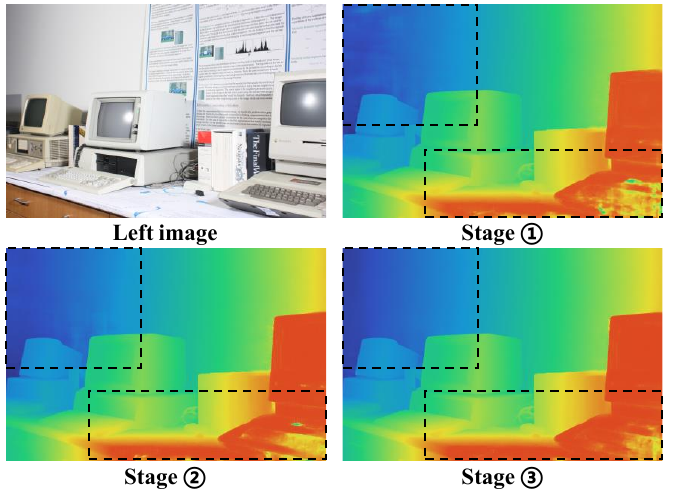}
   \end{center}
   \caption{
    Effects of the proposed three-stage training strategy.
    The predicted disparity maps are progressively improved across stages, with fewer artifacts in the highlighted regions.}
    \label{fig:ablation_compare}
\end{figure}

\subsection{Training Strategy} \label{sec: training strategy}
To achieve strong zero-shot generalization, we train the proposed model with a large-scale collection of high-quality data following a carefully designed three-stage strategy, as illustrated in Fig.~\ref{fig:training_strategy}. We organize the training data into two categories: synthetic annotated data and real-world unlabeled data. Since synthetic datasets provide accurate ground-truth annotations, we first train the model from scratch on synthetic data to establish robust stereo matching capability. Specifically, our synthetic training set consists of SceneFlow~\cite{mayer2016large} (35K), FallingThings~\cite{fallingthings} (30K), FSD~\cite{wen2025foundationstereozeroshotstereomatching} (1.1M), CREStereo~\cite{li2022practical} (0.2M), VKITTI2~\cite{cabon2020vkitti2} (21K), TartanAir~\cite{tartanair2020iros} (0.31M), and Dynamic Replica~\cite{karaev2023dynamicstereo} (0.14M), resulting in approximately 1.8M annotated stereo pairs in total. Although other synthetic datasets are also available, such as IRS~\cite{wang2019irs}, Sintel~\cite{sintel}, Spring~\cite{Mehl2023_Spring}, and InfinigenSV~\cite{jing2024matchstereovideosbidirectional}, we exclude them due to annotation quality issues or significant domain gaps. For example, IRS contains inaccurate disparity annotations for transparent objects, while Spring and InfinigenSV mainly focus on cinematic or natural scenes whose data distributions differ from our target scenarios.

In \textbf{Stage \ding{172}}, the model is trained from scratch in a standard supervised end-to-end manner, without data augmentation. We adopt the commonly used disparity regression loss $\mathcal{L}_{disp}$:
\begin{equation}
    \mathcal{L}_{disp} = smooth_{L_1}(\mathbf{D} - \mathbf{D}_{gt}),
    \label{eq:11}
\end{equation}
where $\mathbf{D}$ and $\mathbf{D}_{gt}$ denote the predicted disparity and the ground-truth disparity, respectively. For the iterative model LAS2-H, we further employ L1 loss on all predicted disparities, where the loss weights are exponentially increased across iterations \cite{lipson2021raft}, formulated as:
\begin{equation}
    \mathcal{L}_{iter} = \mathcal{L}_{disp} + \sum_{k=1}^{n} \gamma^{n-k} \left\| \mathbf{D}_{k} - \mathbf{D}_{gt} \right\|_{1},
    \label{eq:12}
\end{equation}
where $\gamma=0.9$, $\{\mathbf{D}_{k}\}_{k=1}^{n}$ denotes the sequence of disparity predictions over $n$ refinement iterations.

\begin{table*}[t] \addtolength{\tabcolsep}{3pt}
\caption{Zero-shot generalization results on four public benchmarks: KITTI 2012 \cite{kitti12}, KITTI 2015 \cite{kitti15}, ETH3D \cite{eth3d}, and Middlebury (H) \cite{middlebury}. The most commonly used metrics are adopted. Methods are allowed to train on any existing datasets excluding the four target domains. Accurate methods are shown as reference. The weights and parameters are fixed for evaluation. Latency (ms) is measured at the KITTI resolution of $384 \times 1248$. $^{\dag}$ denotes results trained on 30M pseudo-labeled samples using the strategy in \cite{guo2024stereo}. $^{\ddag}$ marks results reported by \cite{wen2025foundationstereozeroshotstereomatching}. The \colorbox{best}{\textbf{best}} and \colorbox{hicell}{second best} are marked with colors.}
\centering
\small
\begin{tabular}{lcccccccccc} 
\toprule
\multirow{2}{*}{Method} &  \multicolumn{2}{c}{KITTI 2012} & \multicolumn{2}{c}{KITTI 2015} & \multicolumn{2}{c}{ETH3D} & \multicolumn{2}{c}{Middlebury} &  \multicolumn{2}{c}{Latency}  \\ 
& D1 & EPE & D1  & EPE & Bad 1.0 & EPE & Bad 2.0 & EPE & H200 & Orin \\ 
\hline
\rowcolor{gray!10} \multicolumn{11}{l}
{\hspace{0pt}\textit{Efficient methods: Feed-forward}}\\
Fast-ACVNet+ \cite{xu2023accurate} & 3.46 & 0.85 & 4.21 & 1.03 & 4.92 & 0.36 & 6.56 & 1.00  & 14.3 & 221  \\
LightStereo-S \cite{guo2024lightstereochannelboostneed} & 4.54 & 1.01 & 5.19 & 1.14 & 8.75 & 0.48 & 12.07 & 1.89 & \second{7.1} & \second{89}  \\
LightStereo-M \cite{guo2024lightstereochannelboostneed} & 4.10 & 0.99 & 4.97 & 1.13 & 5.33 & 0.41 & 10.85 & 1.51 & 8.8 & 119  \\
LightStereo-L \cite{guo2024lightstereochannelboostneed} & 3.70 & 0.88 & 4.69 & 1.07 & 4.44 & 0.35 & 6.74 & \second{0.89}  & 12.9 & 232  \\
BANet-2D \cite{xu2025banet} & {3.90} & 0.93 & {4.71} & {1.07} & 5.92 & 0.38 & 10.05 & 1.34  & 11.6 & 126  \\
BANet-3D \cite{xu2025banet} & 4.50 & 0.99 & 4.13 & 1.03 & 4.82 & 0.36 & 8.10 & 0.99 & 16.5 & 225  \\
StereoAnything-L$^{\dag}$ \cite{guo2024stereo} & 4.00 & {0.92} & 4.81 & 1.10 & {3.81} & {0.31} & {9.82} & {1.21} & 12.9 & 232  \\
LAS \cite{jing2025liteanystereo} & {3.04} & {0.79} & {3.87} & {0.99} & {3.53} & {0.32} & {7.51} & {0.94} & 12.7 & 193  \\
\hdashline
{LAS2-S (ours)} & 2.97 & 0.78 & 3.83 & 0.99 & 3.34 & 0.32 & 8.87 & 1.17 & \bestnum{6.6} & \bestnum{81} \\
% {LAS2-M (ours)} & \second{2.64} & \second{0.72} & \second{3.51} & \bestnum{0.94} & \second{2.30} & \second{0.27} & \second{6.66} & \second{0.89}  & 8.1 & 101 \\
{LAS2-M (ours)} & \second{2.88} & \second{0.74} & \second{3.61} & \second{0.95} & \second{2.59} & \second{0.27} & \second{5.47} & \bestnum{0.77}  & 8.1 & 101 \\
{LAS2-L (ours)} & \bestnum{2.57} & \bestnum{0.71} & \bestnum{3.38} & \bestnum{0.94} & \bestnum{1.83} & \bestnum{0.23} & \bestnum{5.28} & \bestnum{0.77} & 11.4 & 166 \\
\hline
\rowcolor{gray!10} \multicolumn{11}{l}
{\hspace{0pt}\textit{Efficient methods: Iterative}}\\
Lite-CREStereo++ \cite{Jing_2023_ICCV} & 4.09 & 0.91 & 5.33 & 1.12 & 4.92 & 0.63 & 8.94 & 1.46 & \second{23.1} & \second{482} \\
% RT-IGEV++  \cite{} &  &  &  &  \\
RT-MonSter++  \cite{cheng2025monsterunifiedstereomatching} & 2.97 & 0.76 & \second{3.45} & \second{0.94} & \bestnum{1.77} & \bestnum{0.21} & 6.23 & 0.83 & 36.3 & 763 \\
Fast-FoundationStereo  \cite{wen2026fastfoundationstereo} & \second{2.90} & \second{0.75} & 3.66 & \second{0.94} & \second{1.83} & \second{0.22} & \second{3.73} & \second{0.63} & 27.3 & 918 \\
% {LAS2-H (ours)} & \bestnum{2.36} & \bestnum{0.69} & \bestnum{3.07} & \bestnum{0.91} & \bestnum{1.32} & \bestnum{0.20} & \second{4.79} & \second{0.73} & \bestnum{15.1} & \bestnum{344} \\
{LAS2-H (ours)} & \bestnum{2.64} & \bestnum{0.69} & \bestnum{3.31} & \bestnum{0.90} & \second{1.83} & \second{0.22} & \bestnum{3.71} & \bestnum{0.61} & \bestnum{15.1} & \bestnum{344} \\
\hline
\hline
\rowcolor{gray!10} \multicolumn{11}{l}{\hspace{0pt}\textit{Accurate methods}}\\
Selective-IGEV$^{\ddag}$ \cite{wang2024selective} & 3.20 & -- & 4.50 & -- & 3.40 & -- & 7.50 & -- & 129.4 & OOM \\
MonSter++ \cite{cheng2025monsterunifiedstereomatching} & 2.94 & 0.71 & 3.26 & 0.90 & 0.88 & 0.16 & 2.10 & 0.46 & 263.1 & OOM \\
PromptStereo \cite{wang2026promptstereo} & 3.09 & 0.70 & 3.21 & 0.88 & 0.79 & 0.15 & 2.22 & 0.44 & 166.6 & OOM \\
FoundationStereo \cite{wen2025foundationstereozeroshotstereomatching} & 2.51 & 0.67 & 2.83 & 0.86 & 0.49 & 0.14 & 1.12 & 0.37 & 292.2 & OOM \\
\bottomrule
\end{tabular}
% \vspace{-.6em}
\label{tab:comparison_zeroshot}
\end{table*}

In \textbf{Stage \ding{173}}, we introduce a self-distillation strategy to improve feature robustness. Both teacher and student models have the same architectures initialized from the first stage. The teacher model receives clean inputs, while the student model is exposed to strongly perturbed inputs, encouraging domain-invariant representation learning. In addition to the disparity loss, we impose a feature alignment loss $\mathcal{L}_{feat}$:
\begin{equation}
    \mathcal{L}_{feat} = 1 - \frac{1}{HW} \sum_{i=1}^{HW} \cos(F_i, F'_i),
    \label{eq:13}
\end{equation}
where $F_i$ and $F'_i$ are feature vectors from the teacher and student models, respectively. Here, we evaluate several distillation schemes: (a) training only the student model with fixed teacher model weights, (b) updating the teacher model via Exponential Moving Average (EMA) \cite{polyak1992acceleration}, and (c) directly copying student model weights to the teacher model at each iteration. Our ablation study in Section~\ref{sec:Ablation Study} shows that the simplest fixed-teacher strategy achieves the best performance. Therefore, we adopt scheme (a) in the second stage.

High-quality real-world stereo annotations remain scarce and are often sparse, e.g., LiDAR-based ground truth, which limits the scalability of supervised training. In contrast, large-scale unlabeled real-world stereo pairs are much easier to collect, but remain under-explored for efficient zero-shot stereo matching. Therefore, in \textbf{Stage \ding{174}}, we further adapt the lite model to real-world data using 0.5M unlabeled stereo pairs, as summarized in Tab.~\ref{tab:dataset_overview}. We generate dense pseudo labels with FoundationStereo~\cite{wen2025foundationstereozeroshotstereomatching}, a high-capacity stereo foundation model. For datasets that provide sparse annotations, such as DrivingStereo~\cite{yang2019drivingstereo}, we still use dense pseudo labels instead of sparse ground truth for training, since dense supervision provides more complete spatial constraints. The Weather subset is excluded and kept only for evaluation.

Although the teacher model provides strong predictions, some pseudo-label errors are structured and can be explicitly identified.  As shown in Fig.~\ref{fig:training_strategy}, we  apply label filtering before using the raw pseudo labels for supervision. Specifically, we combine three complementary masks. First, we perform a standard left-right consistency check to remove geometrically inconsistent predictions. Given the left and right pseudo disparities $\mathbf{D}_{L}$ and $\mathbf{D}_{R}$, the mask is computed as:
\begin{equation}
M_{LR} =
\mathbf{1}\left(
\left| \mathbf{D}_{L} - \mathcal{W}(\mathbf{D}_{R}, \mathbf{D}_{L}) \right| < \tau_{LR}
\right),
\label{eq:14}
\end{equation}
where $\mathcal{W}(\cdot)$ denotes disparity-based warping, $\tau_{LR}$ is set to $1$ in practice, and $\mathbf{1}(\cdot)$ is an indicator function that outputs $1$ if the condition is satisfied and $0$ otherwise.

\begin{table*}[t] \addtolength{\tabcolsep}{1pt}
\caption{Comparison of model performance on the DrivingStereo Weather subset \cite{yang2019drivingstereo}. All methods use the same checkpoints as those in Tab.~\ref{tab:comparison_zeroshot}.
Lower values indicate better performance for both metrics.}
% \vspace{-.6em}
\centering
\small
\begin{tabular}{lcccccccccccc}
\toprule
\multirow{2}{*}{Method} & \multicolumn{2}{c}{Cloudy} & \multicolumn{2}{c}{Foggy} & \multicolumn{2}{c}{Rainy} & \multicolumn{2}{c}{Sunny} & \multicolumn{2}{c}{Overall} & \multicolumn{2}{c}{Latency}  \\ 
 & D1 & EPE & D1 & EPE & D1 & EPE & D1 & EPE & D1 & EPE & H200 & Orin \\ 
\hline
\rowcolor{gray!10} \multicolumn{13}{l}
{\hspace{0pt}\textit{Efficient methods: Feed-forward}}\\
Fast-ACVNet+~\cite{xu2023accurate} & 6.71 & 2.10 & 18.41 & 3.80 & 41.03 & 9.18 & 5.72 & 1.81 & 17.97 & 4.22 & 14.3 & 221 \\
LightStereo-S \cite{guo2024lightstereochannelboostneed} & 6.32 & 1.71 & 12.39 & 2.12 & 23.34 & 3.58 & 6.03 & 1.78 & 12.02 & 2.30 & \second{7.1} & \second{89}  \\
LightStereo-M \cite{guo2024lightstereochannelboostneed} & 5.18 & 1.68 & 10.61 & 1.96 & 21.57 & 2.73 & 5.73 & 1.74 & 10.77 & 2.03 & 8.8 & 119 \\
LightStereo-L \cite{guo2024lightstereochannelboostneed} & 4.72 & 1.60 & 10.45 & 1.98 & 25.40 & 3.53 & 5.03 & 1.63 & 11.40 & 2.19 & 12.9 & 232 \\
BANet-2D \cite{xu2025banet} & 5.14 & 1.59 & 11.80 & 2.17 & 23.81 & 2.89 & 5.56 & 1.72 & 11.58 & 2.09 & 11.6 & 126 \\
BANet-3D \cite{xu2025banet} & 9.15 & 2.46 & 31.83 & 8.21 & 34.64 & 10.69 & 9.76 & 2.65	& 21.35 & 6.00 & 16.5 & 225 \\
StereoAnything-L \cite{guo2024stereo} & 6.53 & 1.72 & 16.53 & 2.27 & 22.41 & 4.19 & 9.05 & 2.03 & 13.63 & 2.55 & 12.9 & 232 \\
LAS \cite{jing2025liteanystereo} & 3.65 & 1.47 & 6.78 & 1.64 & 20.69 & 2.61 & 3.84 & 1.47 & 8.74 & 1.80 & 12.7 & 193 \\
\hdashline
{LAS2-S (ours)} & \second{2.78} & \bestnum{1.34} & \bestnum{5.51} & \bestnum{1.49} & \bestnum{17.38} & \bestnum{2.38} & \second{3.25} & \second{1.32} & \bestnum{7.23} & \bestnum{1.63} & \bestnum{6.6} & \bestnum{81} \\
% {LAS2-M (ours)} & \bestnum{2.69} & \bestnum{1.33} & 5.76 & \second{1.51} & \second{20.40} & \second{2.59} & \second{3.05} & \bestnum{1.27} & \second{7.98} & \second{1.68} & 8.1 & 101 \\
{LAS2-M (ours)} & {3.24} & {1.42} & 6.37 & {1.60} & \second{17.57} & \second{2.40} & 3.49 & {1.39} & \second{7.67} & \second{1.70} & 8.1 & 101 \\
{LAS2-L (ours)} & \bestnum{2.76} & \second{1.36} & \second{5.62} & \second{1.53} & 20.56 & 2.62 & \bestnum{2.99} & \bestnum{1.29} & 7.98 & \second{1.70} & 11.4 & 166 \\
\hline
\rowcolor{gray!10} \multicolumn{13}{l}
{\hspace{0pt}\textit{Efficient methods: Iterative}}\\
Lite-CREStereo++ \cite{Jing_2023_ICCV} & 5.43 & 1.62 & 9.85 & 1.83 & 24.03 & 2.94 & 4.94 & {1.60} & 11.06 & 2.00 & \second{23.1} & \second{482} \\
% RT-IGEV++  \cite{} &  &  &  &  \\
RT-MonSter++  \cite{cheng2025monsterunifiedstereomatching} & 4.61 & \second{1.47} & 10.94 & 1.87 & \bestnum{13.75} & 2.81 & 4.70 & \second{1.57} & \second{8.50} & 1.93 & 36.3 & 763 \\
Fast-FoundationStereo  \cite{wen2026fastfoundationstereo} & \second{3.86} & 1.54 & \second{7.09} & \second{1.65} & 21.38 & \second{2.80} & \second{4.26} & 1.63 & 9.15 & \second{1.91} & 27.3 & 918 \\
% {LAS2-H (ours)} & \bestnum{2.63} & \bestnum{1.33} & \bestnum{5.91} & \bestnum{1.54} & \second{20.25} & \bestnum{2.59} & \bestnum{2.98} & \bestnum{1.28} & \bestnum{7.94} & \bestnum{1.69} & \bestnum{15.1} & \bestnum{344} \\
{LAS2-H (ours)} & \bestnum{2.97} & \bestnum{1.38} & \bestnum{5.84} & \bestnum{1.56} & \second{17.77} & \bestnum{2.43} & \bestnum{3.16} & \bestnum{1.32} & \bestnum{7.44} & \bestnum{1.67} & \bestnum{15.1} & \bestnum{344} \\
\hline
\hline
\rowcolor{gray!10} \multicolumn{13}{l}{\hspace{0pt}\textit{Accurate methods}}\\
MonSter++ \cite{cheng2025monsterunifiedstereomatching} & 2.88 & 1.27 & 6.77 & 1.49 & 7.45 & 2.96 & 3.72 & 1.37 & 5.21 & 1.77 & 263.1 &  OOM \\
PromptStereo \cite{wang2026promptstereo} & 4.45 & 1.56 & 7.74 & 1.71 & 23.01 & 2.80 & 4.78 & 1.55 & 10.00 & 1.91 & 166.6 & OOM \\
FoundationStereo \cite{wen2025foundationstereozeroshotstereomatching} & 3.85 & 1.53 & 7.67 & 1.82 & 27.01 & 3.96 & 4.31 & 1.57 & 10.71 & 2.22 & 292.2 & OOM \\
\bottomrule
\end{tabular}
% \vspace{-.6em}
\label{tab:drivingstereo-weather}
\end{table*}

Second, we introduce an edge-aware mask to suppress artificial disparity discontinuities. Learning-based stereo teachers may produce sharp disparity variations in textureless regions, even when such variations are not supported by visible image boundaries. To identify these, we compute the gradient magnitudes of the disparity and the left image, and detect strong edges using per-image quantile thresholds. Pixels with strong disparity gradients but no corresponding image evidence are marked as unreliable:
\begin{equation}
M_{edge} =
1 -
\mathbf{1}\left(
\left\| \nabla \mathbf{D}_{L} \right\|_1 > q_{d}
\right)
\cdot
\left(
1 -
\mathbf{1}\left(
\left\| \nabla I_{L} \right\|_1 > q_{I}
\right)
\right),
\label{eq:15}
\end{equation}
where $q_{d}$ and $q_{I}$ are the per-image quantile thresholds for disparity and image gradients. We set both thresholds to the $90$-th percentile in practice and erode the resulting mask with a $3 \times 3$ kernel to remove boundary-adjacent uncertain pixels. Third, we use a segmentation model~\cite{carion2025sam} to identify sky regions and set the mask $M_{{sky}}$ to $0$. The final valid mask $M_{{valid}}$ is obtained by combining the three masks. Only valid pixels are used for supervision.

After removing the unreliable regions, another challenge comes from optimization during the synthetic-to-real transition. At the beginning of Stage \ding{174}, the model is still biased toward synthetic data and may produce relatively large errors on real-world images, even at pixels retained by the valid mask. These high-error pixels  may correspond to difficult regions, occlusion boundaries, or cases where the lite model has not yet adapted to real-world appearance. If optimized directly, a small number of such pixels can dominate the gradients and destabilize training. To mitigate this problem, we adopt an error-clamped disparity loss. Specifically, we truncate the per-pixel disparity loss with an upper bound:
\begin{equation}
\mathcal{L}_{clamp} =
\sum
M_{valid}
\cdot
\min\left(
L_{disp},
\tau_{clamp}
\right),
\label{eq:16}
\end{equation}
where $\tau_{clamp}$ is an empirically estimated maximum allowed value. This simple but effective strategy makes real-world adaptation more stable: the model focuses on the majority of reliable regions instead of being dominated by a few high-error pixels, enabling a smoother transition from synthetic to real-world data.

We further observe that data quality and domain diversity are more important than its raw scale at this stage. Simply adding more stereo pairs does not necessarily improve zero-shot generalization. For example, Stereo4D~\cite{jin2024stereo4d} contains 18M stereo pairs mined from internet videos, but its limited resolution restricts fine-grained supervision. HRWSI~\cite{hrwsi} contains high-resolution stereo images but suffers from rectification artifacts, while domain-specific datasets such as SCOD~\cite{scod} provide limited scene diversity. Therefore, we prioritize well-rectified and diverse real-world stereo pairs in Stage \ding{174}. We do not apply self-distillation in this stage, as it brings no observable gain.  Fig.~\ref{fig:ablation_compare} shows how the proposed three-stage strategy progressively improves disparity estimation, reducing local artifacts and producing cleaner object boundaries.

\begin{figure*}[t]
   \begin{center}
   \includegraphics[width=1\linewidth]{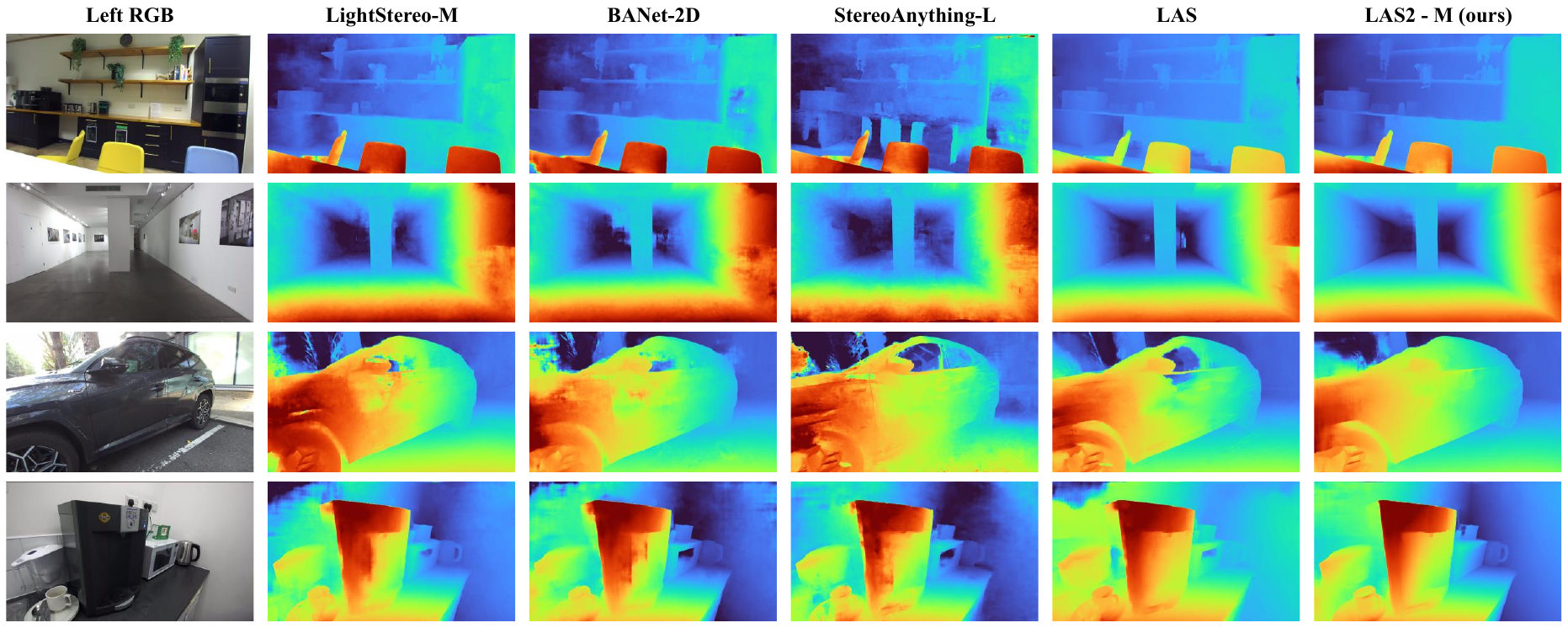}
   \end{center}
    \caption{
    Qualitative comparison with feed-forward stereo methods on in-the-wild stereo images.
    All methods use the same checkpoints as in Tabs.~\ref{tab:comparison_zeroshot} and~\ref{tab:drivingstereo-weather}.
    The examples cover challenging real-world scenes with reflections, complex illumination, low-texture regions, and repetitive patterns.
    Compared with existing methods, LAS2-M produces cleaner object boundaries, fewer texture-copy artifacts, and smoother yet structurally faithful disparity maps, demonstrating stronger zero-shot generalization across diverse scenarios.}
   \label{fig:qualitative comparison}
\end{figure*}

\section{Experiments} \label{sec: Experiments}

\subsection{Benchmarks, Metrics, and Baselines}

\noindent \textbf{Benchmarks.} We evaluate our method on five widely-used real-world stereo datasets. {Middlebury}~\cite{middlebury} is an indoor dataset containing 15 stereo pairs with high-quality ground truth disparities captured using structured light. We report results under the half-resolution and non-occluded evaluation settings. {ETH3D}~\cite{eth3d} consists of 27 grayscale stereo pairs with laser-scanned ground truth, covering both indoor and outdoor scenes. {KITTI 2012}~\cite{kitti12} and {KITTI 2015}~\cite{kitti15} contain 194 and 200 stereo pairs, respectively, captured in outdoor driving environments, with ground truth obtained from LiDAR. The DrivingStereo weather split \cite{yang2019drivingstereo} contains driving scene images under four different weather conditions, with 500 frames for each weather category. We report results on this dataset under the full resolution evaluation setting.

\noindent \textbf{Evaluation Metrics.} For all datasets, we report the average End-Point Error (EPE), which measures the mean per-pixel disparity error. For Middlebury and ETH3D, we additionally report the percentage of pixels whose disparity error exceeds a threshold $X$, denoted as Bad-$X$.  For KITTI and DrivingStereo weather datasets, we report the D1 error, defined as the percentage of pixels whose disparity error is larger than both 3 pixels and 5\% of the ground-truth disparity.

\noindent \textbf{Baselines.} To ensure a fair comparison, we re-evaluate all baseline methods under consistent settings on our local machine. This avoids discrepancies caused by different benchmark configurations, such as occlusion masking and metric definitions, e.g., D1 versus Bad-3.0. Most efficient feed-forward methods release only SceneFlow-pretrained weights; therefore, we retrain their models using the official code and the same synthetic training data as our method. For Fast-FoundationStereo~\cite{wen2026fastfoundationstereo}, 6 variants are released. We adopt the ``20-30-48'' variant with 4 iterations, as it is the fastest configuration. The largest variant cannot be deployed on NVIDIA Orin due to memory constraints.

\begin{figure*}[t]
   \begin{center}
   \includegraphics[width=1\linewidth]{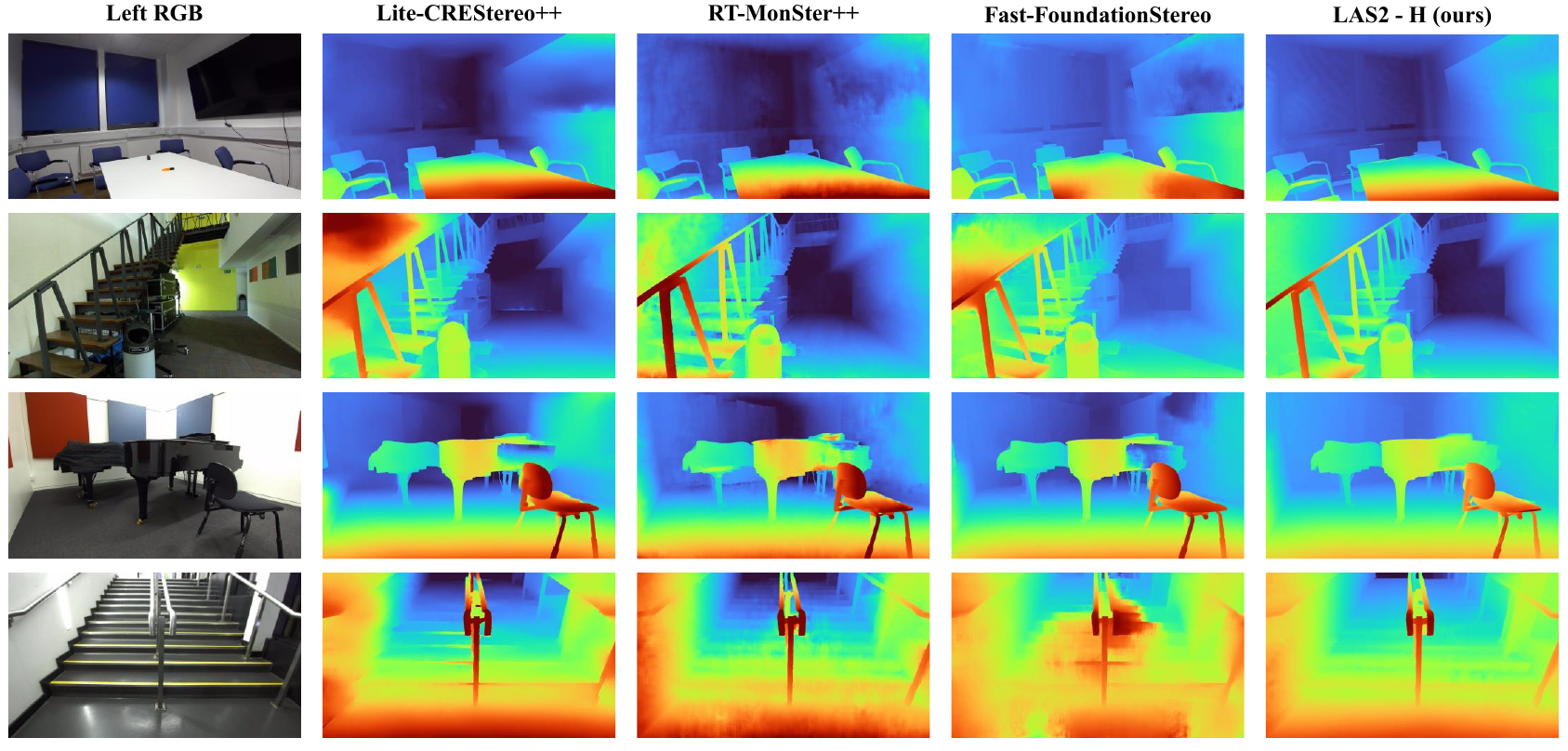}
   \end{center}
    \caption{
    Qualitative comparison with iterative stereo methods on in-the-wild images.
    The examples include challenging indoor scenes with low-texture areas, thin structures, repeated patterns, large depth discontinuities, and complex layouts.
    While existing iterative methods often introduce noisy artifacts, blurred boundaries, or incomplete structures, the proposed LAS2-H produces smoother and more spatially coherent disparity maps with sharper object boundaries.
    These results show that our method improves zero-shot robustness while preserving fine geometric details in challenging real-world scenes.
    }
   \label{fig:qualitative comparison2}
\end{figure*}

\subsection{Implementation Details}

The LAS2 series of models are implemented in PyTorch. We train the model for 200K, 50K, and 200K steps in Stage \ding{172}, Stage \ding{173}, and Stage \ding{174}, respectively, with a total batch size of 128 on NVIDIA H200 GPUs. For LAS2-H, we initialize the feature extraction and cost aggregation module with the weights of LAS2-M and freeze it during the first 100K steps of Stage \ding{172}. We then fine-tune all model parameters in the remaining training steps. We adopt the AdamW optimizer~\cite{adam} with a one-cycle learning rate schedule, where the peak learning rate is set to $2 \times 10^{-4}$. During training, input images are randomly cropped to $384 \times 768$. The maximum disparity value, $\text{D}_{\text{max}}$, is set to 192, following the configuration used in prior works~\cite{guo2024lightstereochannelboostneed,xu2025banet}. In Stage \ding{173}, we apply strong perturbations to the training pairs. These include random color jitter with large variations in brightness, contrast, saturation, and hue; optional $5{\times}5$ Gaussian blur with $\sigma \in [0.1, 2.0]$; and random gamma correction with $\gamma \in [0.7, 1.5]$. These transformations are applied either symmetrically to both views or asymmetrically with a small probability.

\subsection{Evaluation}
\noindent \textbf{Zero-shot Generalization.}
Tab.~\ref{tab:comparison_zeroshot} reports the zero-shot generalization results on four public benchmarks, including KITTI 2012, KITTI 2015, ETH3D, and Middlebury.
All methods are evaluated with fixed checkpoints and are not trained on the target domains.
Among feed-forward efficient methods, our LAS2 series consistently achieves the best performance.
Compared with LAS~\cite{jing2025liteanystereo}, LAS2-M improves all reported accuracy metrics while reducing the latency. Our smallest variant, LAS2-S, already achieves competitive zero-shot accuracy while obtaining the lowest latency in the feed-forward group, demonstrating the effectiveness of the lightweight design.
By increasing the model capacity, LAS2-L further improves accuracy and achieves the best overall performance among feed-forward efficient methods.

For efficient iterative methods, LAS2-H achieves the strongest overall performance across the four benchmarks while remaining substantially faster than existing iterative baselines. Compared with Fast-FoundationStereo~\cite{wen2026fastfoundationstereo}, LAS2-H achieves better accuracy on KITTI and Middlebury, matches it on ETH3D, and substantially reduces the latency on both H200 and Orin. This shows that initializing the iterative refinement framework with our feed-forward model is an effective way to improve accuracy while keeping the computation practical. High-accuracy reference methods such as FoundationStereo~\cite{wen2025foundationstereozeroshotstereomatching} and MonSter++~\cite{cheng2025monsterunifiedstereomatching} also achieve strong zero-shot results, but require much larger computational budgets and run out of memory on the edge device in this setting. In contrast, our models remain deployable on Orin, with LAS2-H even outperforming several high-accuracy references on KITTI while using substantially lower latency.

\noindent \textbf{Challenging Weather.}
Tab.~\ref{tab:drivingstereo-weather} evaluates model robustness on the DrivingStereo weather subset. This benchmark covers diverse weather conditions, including cloudy, foggy, rainy, and sunny scenes. Among feed-forward efficient methods, LAS2-S achieves the best overall D1 and EPE while also obtaining the lowest latency, showing that the lightweight variant is already highly robust under challenging weather. LAS2-M and LAS2-L also outperform previous feed-forward efficient baselines on the overall metrics, confirming that the proposed architecture and training strategy generalize well beyond standard benchmarks.

Among efficient iterative methods, LAS2-H achieves the best overall performance while being considerably faster than other iterative baselines. Notably, although FoundationStereo is only used as the pseudo-label teacher during training, LAS2-H outperforms both Fast-FoundationStereo and FoundationStereo on the overall metrics of this benchmark. These results suggest that the proposed pseudo-label filtering and staged training strategy can effectively transfer useful geometric priors from the teacher while improving robustness and efficiency under challenging weather conditions.

\begin{table*}[t] \addtolength{\tabcolsep}{3pt}
\caption{Results on the {KITTI 2012}~\cite{kitti12} and {KITTI 2015}~\cite{kitti15} leaderboard. Latency is measured under the same setting on NVIDIA H200 and Orin NX 8G. Among efficient stereo methods, our finetuned LAS2-M achieves the best results on KITTI 2015 while also delivering the fastest inference speed, ranking first on the leaderboard at the time of submission.}
% \vspace{-.6em}
\centering
\small
\begin{tabular}{lccccc|ccccc}
\toprule
\multirow{2}{*}{Method} & \multicolumn{5}{c|}{{KITTI 2012}} & \multicolumn{3}{c}{{KITTI 2015}} & \multicolumn{2}{c}{Latency}  \\ 
& 3-noc & 3-all & 4-noc & 4-all & EPE-noc / all & D1-bg & D1-fg & D1-all & H200 & Orin \\ 
\midrule
% \rowcolor{gray!10} \multicolumn{11}{l}
% {\hspace{0pt}\textit{Efficient methods: Feed-forward}}\\
AANet+~\cite{xu2020aanet} & 1.55 & 2.04 & 1.20 & 1.58 & {0.4 / 0.5} & 1.65 & 3.96 & 2.03 & -- & -- \\
% DecNet~\cite{yao2021decomposition} & - & - & - & - & - & 2.07 & 3.87 & 2.37 & - \\
BGNet+~\cite{xu2021bilateral} & 1.62 & 2.03 & 1.16 & 1.48 & 0.5 / 0.6 & 1.81 & 4.09 & 2.19 & -- & -- \\
HITNet~\cite{tankovich2021hitnet} & 1.41 & 1.89 & 1.14 & 1.53 & {0.4 / 0.5} & 1.74 & 3.20 & 1.98 & -- & -- \\
CoEx~\cite{bangunharcana2021correlate} & 1.55 & 1.93 & 1.15 & 1.42 & 0.5 / 0.5 & 1.79 & 3.82 & 2.13 & -- & -- \\
% MobileStereoNet-2D~\cite{shamsafar2022mobilestereonet} & - & - & - & - & - & 2.49 & 4.53 & 2.83 & -- & -- \\
% MobileStereoNet-3D~\cite{shamsafar2022mobilestereonet} & - & - & - & - & - & 2.75 & 3.87 & 2.10 & -- & -- \\
% Fast-ACVNet~\cite{xu2022acvnet} & 1.68 & 2.13 & 1.23 & 1.56 & 0.5 / 0.6 & 1.82 & 3.93 & 2.17 & -- & -- \\
Fast-ACVNet+~\cite{xu2023accurate} & 1.45 & 1.85 & 1.06 & 1.36 & 0.5 / 0.5 & 1.70 & 3.53 & 2.01 & 14.3 & 221 \\
% LightStereo-S \cite{guo2024lightstereochannelboostneed} & 1.88 & 2.34 & 1.30 & 1.65 & 0.6 / 0.6 & 2.00 & 3.80 & 2.30 & 7.1 & 89 \\
LightStereo-M \cite{guo2024lightstereochannelboostneed} & 1.56 & 1.91 & 1.10 & 1.36 & 0.5 / 0.5 & 1.81 & 3.22 & 2.04 & \second{8.8} & \second{119} \\
LightStereo-L \cite{guo2024lightstereochannelboostneed} & 1.55 & 1.87 & 1.10 & 1.33 & 0.5 / 0.5 & 1.78 & \bestnum{2.64} & 1.93 & 12.9 & 232 \\
BANet-2D \cite{xu2025banet} & 1.38 & 1.79 & 1.01 & 1.32 & 0.5 / 0.5 & {1.59} & {3.03} & {1.83} & 11.6 & 126 \\
BANet-3D \cite{xu2025banet} & {1.27} & {1.72} & {0.95} & {1.27} & 0.5 / 0.5 & {1.52} & {3.02} & {1.77} & 16.5 & 225 \\
% \hdashline
% {LAS2-S (ours)} &  &  &  &  &  &  &  &  & 6.6 & 81 \\
% {LAS2-M (ours)} &  &  &  &  &  &  &  &  & 8.1 & 101 \\
% {LAS2-L (ours)} &  &  &  &  &  &  &  &  & 11.4 & 166 \\
% \midrule
% \rowcolor{gray!10} \multicolumn{11}{l}
% {\hspace{0pt}\textit{Efficient methods:Iterative-based}}\\
Lite-CREStereo++ \cite{Jing_2023_ICCV} & 1.43 & 1.82 & 1.12 & 1.44 & 0.5 / 0.5 & 1.79 & 3.53 & 2.08 & 23.1 & 482 \\
RT-MonSter++  \cite{cheng2025monsterunifiedstereomatching}  & \bestnum{1.07} & \bestnum{1.41} & \second{0.80} & \second{1.05} & 0.4 / 0.4 & 1.47 & \second{2.78} & \second{1.69} & 36.3 & 763 \\
% {LAS2-H (ours)} &  &  &  &  &  &  &  &  & 15.1 & 344 \\
LAS \cite{jing2025liteanystereo} & \second{1.09} & \second{1.49} & \bestnum{0.76} & \bestnum{1.04} & {0.4 / 0.5} & \second{1.36}	& {3.45} & {1.71} & 12.7 & 193 \\
{LAS2-M (ours)} & 1.13 & 1.51 & 0.81 & 1.06 & 0.5 / 0.5 & \bestnum{1.33} & 3.00 & \bestnum{1.61} & \bestnum{8.1} & \bestnum{101} \\
\bottomrule
\end{tabular}
% \vspace{-1.3em}
\label{tab:kitti_benchmark}
\end{table*}

\begin{table*}[t] \addtolength{\tabcolsep}{5pt}
\caption{Ablation study on KITTI 2012 \cite{kitti12} (K.12), KITTI 2015 \cite{kitti15} (K.15), ETH3D \cite{eth3d} (E.), and Middlebury \cite{middlebury} (M.). We report D1 for KITTI, Bad 1.0 for ETH3D, and Bad 2.0 for Middlebury. We also report MACs and latency on NVIDIA Orin NX 8G. Models are trained based on LAS2-M for 150K iterations without data augmentation on a 1.4M-image subset from synthetic datasets \cite{wen2025foundationstereozeroshotstereomatching, li2022practical, cabon2020vkitti2, fallingthings, mayer2016large} using the default operations in \cite{xu2025banet}. The final default settings are underlined.}
\centering
\small
\begin{tabular}{cccccccc} 
\toprule
Module & Block &  K.12 & K.15 & E. & M. & MACs (G) & Latency (ms) \\
\hline
\multirow{6}{*}{Cost Aggregation}  & ConvNeXt \cite{liu2022convnet} & 4.92 & 4.86 & 7.99 & 9.76 & \textbf{31.2} & 162 \\
& MobileNet V2 \cite{sandler2018mobilenetv2} & 4.81 & 4.85 & 6.25 & 10.73 & 32.2 & 118 \\
& MobileNet V3 \cite{howard2019searching} & 4.54 & 4.85 & \textbf{5.59} & 10.41 & 32.2 & 145 \\
& EfficientNet V2 \cite{tan2021efficientnetv2} & 5.62 & \textbf{4.52} & 8.76 & 10.28 & 58.4 & 123 \\
& \underline{FasterNet \cite{chen2023run}} & \textbf{4.49} & 4.57 & 5.62 & \textbf{9.49} & {33.9} & \textbf{107} \\
& GhostNet \cite{han2020ghostnet} & 4.82 & 4.93 & 5.64 & 11.53 & 32.9 & 185 \\
\hline
\multirow{3}{*}{Feature Extraction}  & MobileNet V2 \cite{sandler2018mobilenetv2} & \textbf{4.49} & \textbf{4.57} & 5.62 & \textbf{9.49} & \textbf{33.9} & 107 \\
& FasterNet + 1x1 conv \cite{chen2023run} & 5.18 & 4.91 & \textbf{5.16} & 10.98 & 35.6 & \textbf{95} \\
& \underline{FasterNet \cite{chen2023run}} & 4.84 & 4.77 & 5.37 & 10.49 & 47.6 & 101 \\
\bottomrule
\end{tabular}
\label{tab:ablation_backbone}
\end{table*}

\begin{table*}[t] \addtolength{\tabcolsep}{10pt}
\caption{Ablation study on Stage \ding{174} of LAS2-M on  KITTI 2012 \cite{kitti12} (K.12), KITTI 2015 \cite{kitti15} (K.15), ETH3D \cite{eth3d} (E.), and Middlebury \cite{middlebury} (M.). The default settings of the final model are underlined.}
\centering
\small
\begin{tabular}{cccccc} 
\toprule
Case & Settings &  K.12 & K.15 & E. & M. \\
\hline
\multirow{5}{*}{Valid Mask}  & w/o & 2.91 & 3.66 & 3.01 & 5.88  \\
& $M_{LR}$ & 2.91 & 3.73 & 2.76 & \textbf{5.29}  \\
& $M_{LR}$ + $M_{sky}$ & \textbf{2.86} & 3.62 & 2.62 & 5.68  \\
& \underline{$M_{LR}$ + $M_{sky}$ + $M_{edge}$} & 2.88 & \textbf{3.61} & \textbf{2.59} & 5.47  \\
& $M_{LR}$ + $M_{sky}$ + $M_{edge}$ + $M_{rgb}$ & 2.89 & 3.63 & 2.85 & 5.41  \\
\hline
\multirow{4}{*}{Error Clamp}  & w/o & 3.17 & 3.92 & 3.22 & 6.21  \\
& $\tau_{clamp}=5$ & 2.89 & \textbf{3.59} & \textbf{2.54} & 5.57  \\
& \underline{$\tau_{clamp}=10$} & \textbf{2.88} & 3.61 & 2.59 & \textbf{5.47}  \\
& $\tau_{clamp}=20$ & 2.98 & 3.70 & 2.88 & 5.98  \\
\hline
\multirow{2}{*}{Feature Alignment}  & w. & 2.93 & 3.65 & 2.66 & 5.99  \\
& \underline{w/o} & \textbf{2.88} & \textbf{3.61} & \textbf{2.59} & \textbf{5.47} \\
\hline
\multirow{2}{*}{Teacher Model}  & \underline{FS \cite{wen2025foundationstereozeroshotstereomatching}} & \textbf{2.88} & 3.61 & 2.59 & \textbf{5.47}  \\
& FS \cite{wen2025foundationstereozeroshotstereomatching} + MS \cite{cheng2025monsterunifiedstereomatching} + S2M2 \cite{min2025s2m2} & 2.94 & \textbf{3.35} & \textbf{2.10} & 7.53  \\
\hline
\multirow{3}{*}{Extra Data}  & \underline{base set} & \textbf{2.88} & \textbf{3.61} & \textbf{2.59} & {\textbf{5.47}} \\
& base set + Stereo4D \cite{jin2024stereo4d} (1.4M) & 2.93 & 3.74 & 2.94 & 6.52  \\
& base set + Xperience \cite{xperience_10m} (3.6M) & 4.37 & 5.08 & 2.77 & 9.17  \\
\bottomrule
\end{tabular}
\label{tab:ablation_strategy}
\end{table*}

\noindent \textbf{Qualitative Results.}
Figs.~\ref{fig:qualitative comparison} and~\ref{fig:qualitative comparison2} provide qualitative comparisons on in-the-wild stereo images. In Fig.~\ref{fig:qualitative comparison}, existing feed-forward methods often produce less stable disparity maps in challenging scenes. For example, in the long corridor scene, several baselines generate noisy or distorted predictions around the distant wall and floor regions, while in the car scene, reflective surfaces lead to local artifacts on the vehicle body. In contrast, LAS2-M produces smoother and more structurally consistent disparity maps with cleaner object boundaries.

In Fig.~\ref{fig:qualitative comparison2}, existing iterative models struggle with thin structures and large depth discontinuities. In the stair scenes, several baselines produce broken or noisy disparity patterns around the railings and central handrail. LAS2-H produces more spatially coherent disparity while preserving fine geometric details and object boundaries. These qualitative results are consistent with the quantitative comparisons and further demonstrate the robustness of our method on challenging real-world scenes.

\noindent \textbf{In-domain.}
Although in-domain performance is not the primary focus of this work, we also evaluate our model on the KITTI online leaderboard (test set) trained on \cite{kittidepth} without using its original annotations. As shown in Tab.~\ref{tab:kitti_benchmark}, our model achieves the highest accuracy on KITTI 2015 among all published efficient methods.

\begin{table}[t] \small\addtolength{\tabcolsep}{7pt}
\caption{Effects of self distillation in Stage \ding{173} on 1.4M synthetic images.}
\centering
\begin{tabular}{c|cccc}
\toprule
case & K.12 & K.15 & E. & M.  \\
\hline
none  & 4.84 & \textbf{4.77} & 5.37 & 10.49 \\
data aug. & 4.15 & 4.97 & 5.94 & 9.06  \\
\underline{self. dis.} & \textbf{3.85} & {4.78} & \textbf{4.89} & \textbf{8.83}\\
\bottomrule
\end{tabular}
\label{tab:ablation stage2-1}
\end{table}

\begin{table}[t] \small\addtolength{\tabcolsep}{7pt}
\caption{Self distillation choices in Stage~\ding{173} on 1.4M synthetic images.}
\centering
\begin{tabular}{c|cccc}
\toprule
case & K.12 & K.15 & E. & M.  \\
\hline
EMA & 4.46 & 5.25 & 6.84 & 9.64  \\
hard copy  & 4.01 & 4.85 & 6.66 & \textbf{8.83} \\
\underline{fixed} & \textbf{3.85} & \textbf{4.78} & \textbf{4.89} & \textbf{8.83}\\
\bottomrule
\end{tabular}
\label{tab:ablation stage2-2}
\end{table}

\begin{table}[t] \small\addtolength{\tabcolsep}{7pt}
\caption{Effects of the three-stage training strategy on LAS2-M using the full training set.}
\centering
\begin{tabular}{c|cccc}
\toprule
case & K.12 & K.15 & E. & M.  \\
\hline
stage  \ding{172}  & 4.21 & 4.66 & 4.25 & 7.95 \\
stage  \ding{173}  & 3.59 & 4.65 & 4.67 & 6.91  \\
stage  \ding{174}  & \textbf{2.88} & \textbf{3.61} & \textbf{2.59} & {\textbf{5.47}}\\
\bottomrule
\end{tabular}
\label{tab:ablation_3stage}
\end{table}

\begin{table}[t] \small\addtolength{\tabcolsep}{7pt}
\caption{Performance of the proposed training strategy. We apply the same strategy to LightStereo-M and BANet-2D. The results below the dashed line are obtained using the strategy from the previous version~\cite{jing2025liteanystereo}.}
% \vspace{-.6em}
\centering
% \vspace{1mm}
\begin{tabular}{lcccc}
\toprule
{Method} & K.12 & K.15 & E. & M. \\
\midrule
\rowcolor{gray!10} \multicolumn{5}{l}{\hspace{0pt}\textit{LightStereo-M} \cite{guo2024lightstereochannelboostneed}}\\
Stage \ding{172}       & 4.34 & 5.27 & 6.68 & 10.29  \\
Stage \ding{173}       & 3.80 & 4.62 & 5.44 & {8.96} \\
Stage \ding{174}       & \textbf{2.74}  & \textbf{3.74}  & \textbf{2.73}  & \textbf{7.08}  \\
\hdashline
Stage \ding{174} \cite{jing2025liteanystereo}   & 3.35  & 4.14  & 4.22  & 9.85  \\
\midrule
\rowcolor{gray!10} \multicolumn{5}{l}{\hspace{0pt}\textit{BANet-2D} \cite{xu2025banet}}\\
Stage \ding{172}       & 4.34 & 4.78 & 7.71 & 10.54  \\
Stage \ding{173}       & 3.87  & 4.80  & 4.86  & {9.54}  \\
Stage \ding{174}       & \textbf{2.69} & \textbf{3.66} & \textbf{2.61} & \textbf{7.81}  \\
\hdashline
Stage \ding{174} \cite{jing2025liteanystereo}  & 3.28  & 4.08 & 4.05 & 10.30  \\
\bottomrule
\end{tabular}
% \vspace{-1.2em}
\label{tab:universal training strategy}
\end{table}

\begin{table*}[t] \addtolength{\tabcolsep}{9pt}
\caption{
Latency comparison. We report the inference time in milliseconds of different methods on desktop/server GPUs and the embedded Orin NX 8G under different power modes. All methods are evaluated locally using the same input size of $384 \times 1248$ and the same benchmarking protocol. For fair comparison, \texttt{torch.compile} is disabled for all methods.
}

\centering
\small
\begin{tabular}{l|cccc|ccc} 
\toprule
\multirow{2}{*}{Methods} & \multicolumn{4}{c|}{GPU} & \multicolumn{3}{c}{NVIDIA Orin NX 8G} \\
  & RTX 4090 & A5000	& A100 & H200 &  10W & 20W & MAXN \\
\hline
\rowcolor{gray!10} \multicolumn{8}{l}
{\hspace{0pt}\textit{Efficient methods: Feed-forward}}\\
Fast-ACVNet+~\cite{xu2023accurate} & 23.3 & 31.5 & 56.6 &  14.3 &  469 & 380 & 221 \\
LightStereo-S \cite{guo2024lightstereochannelboostneed} & \second{16.2} & \second{19.9} & \second{23.6} & \second{7.1} & 229 & \second{155} & \second{89} \\
LightStereo-M \cite{guo2024lightstereochannelboostneed} & 21.4 & 26.2 & 30.0 & 8.8 & 269 & 205 & 119\\
LightStereo-L \cite{guo2024lightstereochannelboostneed} & 28.9 & 34.7 & 46.8 &	12.9 & 523 & 391 & 232 \\
BANet-2D \cite{xu2025banet} & 30.4 & 35.8 & 40.3 & 11.6 & 294 & 215 & 126 \\
BANet-3D \cite{xu2025banet} & 24.5 & 33.1 & 64.2 & 16.5 & 507 & 393 & 225 \\
StereoAnything-L \cite{guo2024stereo} & 28.9 & 34.7 & 46.8 & 12.9 & 523 & 391 & 232 \\
LAS \cite{jing2025liteanystereo} & 21.4 & 26.7  & 46.6 & 12.7 & 469 & 345 & 193 \\
\hdashline
{LAS2-S (ours)} & \bestnum{11.5} & \bestnum{16.3} & \bestnum{22.9} & \bestnum{6.6} & \bestnum{181} & \bestnum{144} & \bestnum{81} \\
{LAS2-M (ours)} & 16.8 & 21.4 & 29.2 & 8.1 & \second{225} & 179 & 101 \\
{LAS2-L (ours)} & 23.2 & 29.2 & 41.8 & 11.4 & 372 & 283 & 166 \\
\hline
\rowcolor{gray!10} \multicolumn{8}{l}
{\hspace{0pt}\textit{Efficient methods: Iterative}}\\
Lite-CREStereo++ \cite{Jing_2023_ICCV} & 37.8 & \second{55.8} & \second{96.6}  & \second{23.1} & \second{1033} & \second{835} & 482 \\
% RT-IGEV++  \cite{} &  &  &  &  \\
RT-MonSter++  \cite{cheng2025monsterunifiedstereomatching} & 36.8 & 66.5 & 151.3 & 36.3 & 1704 & 1404 & \second{763} \\
Fast-FoundationStereo  \cite{wen2026fastfoundationstereo} & \second{31.7} & 67.0 & 123.0 & 27.3 & 1731 & 1646 & 918 \\
{LAS2-H (ours)} & \bestnum{26.2} & \bestnum{37.2} & \bestnum{65.9} & \bestnum{15.1} & \bestnum{689} & \bestnum{594} & \bestnum{344} \\
\bottomrule
\end{tabular}
\label{tab:latency}
\end{table*}

\subsection{Ablation Study} \label{sec:Ablation Study}
In Tabs.~\ref{tab:ablation_backbone}, \ref{tab:ablation_strategy}, \ref{tab:ablation stage2-1}, \ref{tab:ablation stage2-2}, and \ref{tab:ablation_3stage}, we investigate various design choices and training strategies of our model. Unless otherwise specified, LAS2-M is used as the backbone, and all variants are trained for 150K iterations without data augmentation on a 1.4M-image subset of synthetic datasets \cite{wen2025foundationstereozeroshotstereomatching, li2022practical, cabon2020vkitti2, fallingthings, mayer2016large}. We report D1 on KITTI 2012 \cite{kitti12} (K.12) and KITTI 2015 \cite{kitti15} (K.15), Bad 1.0 on ETH3D \cite{eth3d} (E.), and Bad 2.0 on Middlebury \cite{middlebury} (M.). For architecture ablations, we also report MACs and latency on NVIDIA Orin 8G.

\noindent \textbf{Architecture Design.} Tab.~\ref{tab:ablation_backbone} ablates the block choices for cost aggregation and feature extraction under the same controlled training setting. For cost aggregation, FasterNet~\cite{chen2023run} provides the best accuracy-efficiency trade-off. It achieves the lowest errors on KITTI 2012 and Middlebury, competitive results on KITTI 2015 and ETH3D, and the lowest latency among all cost aggregation variants on Orin. Although ConvNeXt~\cite{liu2022convnet} has slightly lower MACs, its latency is substantially higher, indicating that MACs alone are not sufficient to reflect practical deployment efficiency. EfficientNet V2~\cite{tan2021efficientnetv2} and MobileNet V3~\cite{howard2019searching} perform best on KITTI 2015 and ETH3D, respectively, but their gains are not consistent across datasets. Thus, we adopt FasterNet as the cost aggregation block.

For feature extraction, MobileNet V2 yields superior accuracy, but it has a higher measured latency. We therefore explore FasterNet-based alternatives for more efficient feature extraction. A straightforward design is to use FasterNet followed by a $1{\times}1$ convolution to align its output channel dimension with that of MobileNet V2. This design reduces the latency from 107 ms to 95 ms, but it also leads to noticeable accuracy degradation on KITTI and Middlebury. We further remove the additional $1{\times}1$ convolution and retain the native output channel dimension of FasterNet. This simplified design recovers much of the accuracy loss while still reducing the latency compared with MobileNet V2. Therefore, we use FasterNet for feature extraction in the final model, as it offers a balanced performance with a simpler architecture.

\noindent \textbf{Training Strategy Choices.}
Tab.~\ref{tab:ablation_strategy} studies the key design choices in Stage~\ding{174}. For pseudo-label filtering, the left-right consistency mask brings the largest overall improvement, showing that removing geometrically inconsistent predictions is critical when using dense pseudo labels.
Adding the sky mask gives mixed changes on the reported benchmarks, likely because these test sets contain limited sky regions. Nevertheless, we keep it in the final model for practical real-world deployment, where sky regions are common and often have ambiguous stereo correspondence. The proposed edge mask further improves the overall balance by suppressing unreliable pseudo labels near disparity discontinuities that are not supported by image evidence.
Adding the RGB consistency mask slightly improves Middlebury but degrades the other benchmarks, so it is not used in the final setting.

For error clamping, removing it clearly degrades all metrics, confirming its importance for stable synthetic-to-real adaptation.
Once clamping is applied, $\tau_{\mathrm{clamp}}=5$ and $\tau_{\mathrm{clamp}}=10$ achieve similar strong performance, while a looser threshold of $\tau_{\mathrm{clamp}}=20$ is less effective.
This suggests that the main benefit comes from preventing a few high-error pseudo labels from dominating the gradients.
We choose $\tau_{\mathrm{clamp}}=10$ as a balanced default, which achieves strong overall accuracy.

In addition, we also evaluate several more expensive alternatives.
Feature alignment with real-world data has been shown to be effective in~\cite{wen2026fastfoundationstereo}, but it brings no clear gain in our setting.
Since it requires additional forward passes and substantially increases the training cost, we remove it from the final stage.
Similarly, using multiple teachers improves KITTI 2015 and ETH3D but degrades KITTI 2012 and Middlebury, leading to inconsistent overall performance while increasing pseudo-label generation cost.
Finally, adding more data from Stereo4D~\cite{jin2024stereo4d} or Xperience~\cite{xperience_10m} does not improve performance.
This suggests that simply scaling the amount of data is insufficient when the additional data may suffer from limited resolution, domain bias, or imperfect stereo quality; data quality and diversity are more important than raw quantity.

\noindent \textbf{Effectiveness of the Training Strategy.}
Tabs.~\ref{tab:ablation stage2-1} and~\ref{tab:ablation stage2-2} validate the effect of knowledge distillation in Stage~\ding{173}.
Compared with direct data augmentation, knowledge distillation achieves better overall performance, and using fixed teacher weights gives the most stable results. Tab.~\ref{tab:universal training strategy} further shows that Stage~\ding{174} brings the largest improvement, confirming that synthetic pretraining, self-distillation, and real-world pseudo-label training play complementary roles.

To verify the generality of the proposed strategy, we apply it to LightStereo-M~\cite{guo2024lightstereochannelboostneed} and BANet-2D~\cite{xu2025banet}.
As shown in Tab.~\ref{tab:universal training strategy}, both models benefit substantially from the full training pipeline and clearly outperform the strategy used in the previous version~\cite{jing2025liteanystereo}.
These results show that the proposed training strategy is not tied to LAS2, but can serve as a general recipe for improving efficient zero-shot stereo models.

\subsection{Latency Analysis}
\label{sec:latency_analysis}

Prior works often report inference latency on different hardware platforms and with different benchmarking protocols, making direct speed comparisons unreliable. Moreover, some reported numbers are obtained with implementation-specific acceleration. For a fair comparison, we disable \texttt{torch.compile} for all methods under the same input size of $384 \times 1248$ and the same evaluation protocol. As shown in Tab.~\ref{tab:latency}, LAS2 achieves favorable latency across both desktop/server GPUs and the embedded platform.

\begin{figure}[t]
   \begin{center}
   \includegraphics[width=1\linewidth]{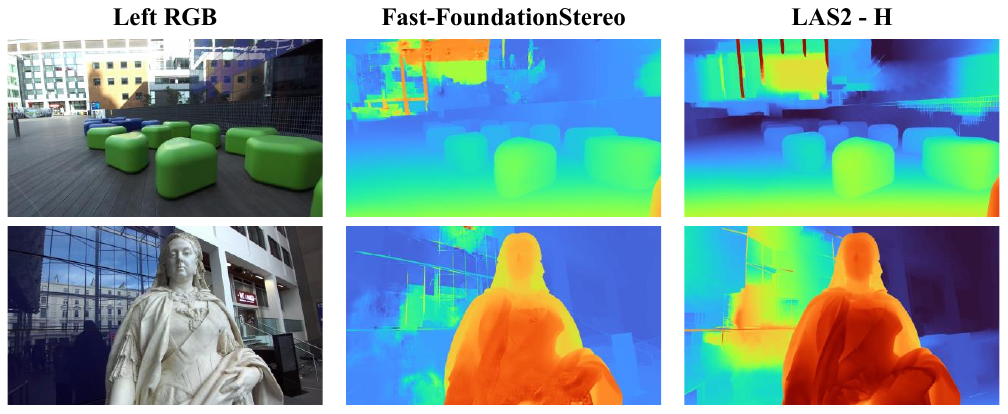}
   \end{center}
    \caption{
    Visualization of failure cases.
    Both the existing method and our method struggle in extremely challenging scenes with severe illumination changes, reflective/transparent surfaces, and ambiguous geometry.
    These examples suggest that such complex conditions remain an open challenge.
    }
    \label{fig:failure_cases}
\end{figure}

For feed-forward stereo methods, LAS2 provides clear efficiency advantages at comparable model scales.
Compared with LightStereo, LAS2-S/M/L are consistently faster than their corresponding small, medium, and large variants across all tested GPUs and Orin power modes.
Compared with the previous LAS model, LAS2-M also substantially reduces latency on all platforms, demonstrating that replacing the expensive 3D aggregation with the proposed 2D aggregation directly improves inference efficiency.
Meanwhile, LAS2-S achieves the lowest latency among all feed-forward methods, making it suitable for latency-sensitive deployment scenarios.

For iterative stereo methods, LAS2-H is consistently faster than recent baselines such as RT-MonSter++ and Fast-FoundationStereo.
The advantage is particularly clear on the embedded Orin platform, where existing iterative methods suffer from substantially higher latency.
This shows that LAS2-H retains the benefit of iterative refinement while greatly reducing its computational overhead.
Together with the accuracy results in previous sections, these results demonstrate that LAS2 offers a favorable speed--accuracy trade-off and strong potential for resource-constrained deployment.

\section{Limitations and Discussion} \label{sec: Limitations}
\label{sec:limitations}

Although LAS2 improves zero-shot generalization while maintaining high efficiency, it still has several limitations.
First, there remains a performance gap between LAS2 and prior-based high-accuracy methods.
These methods benefit from strong monocular depth priors and larger foundation backbones, which provide stronger semantic and geometric representations.
Second, LAS2 is still constrained by the limited scale of high-quality real-world stereo data.
Although our pseudo-label filtering and staged training strategy can effectively exploit unlabeled real-world stereo images, the amount of diverse and high-quality real-world stereo data is still insufficient compared with the scale of data available for monocular foundation models.
This data bottleneck limits the upper bound of current efficient stereo models and motivates future efforts on larger-scale real-world stereo data collection.
Finally, as illustrated in Fig.~\ref{fig:failure_cases}, LAS2, like existing state-of-the-art methods, can still fail in extremely challenging scenes with strong reflections, transparent surfaces, severe illumination changes, or ambiguous geometry.
These cases remain difficult for current stereo matching methods and suggest important directions for future research.

\section{Conclusion} \label{sec: Conclusion}
We present LAS2, an efficient stereo matching model series designed for zero-shot generalization. By revisiting stereo architecture design from the perspective of practical deployment, LAS2 adopts a pure 2D cost aggregation framework that substantially improves measured inference latency on both GPUs and edge devices. Together with a three-stage training strategy that combines synthetic supervision, self-distillation, and real-world knowledge distillation, LAS2 achieves strong generalization across diverse real-world scenarios. We further introduce pseudo-label filtering and error clamping to improve the reliability of real-world pseudo-label supervision and enable smoother synthetic-to-real transfer. Extensive experiments demonstrate that LAS2 achieves state-of-the-art accuracy among efficient stereo methods while maintaining significantly lower latency, narrowing the gap between lightweight and accuracy-oriented stereo models. These results suggest that LAS2 can serve as a practical and deployable solution for efficient stereo matching and provide useful insights for deploying stereo models on real-world hardware.

\noindent \textbf{Acknowledgment.} S. Zafeiriou was funded by the EPSRC Fellowship DEFORM (EP/S010203/1), EPSRC Project GNOMON (EP/X011364/1) and Turing AI Fellowship (EP/Z534699/1). J. Deng was supported by the
NVIDIA Academic Grant. The authors acknowledge the use of resources provided by the Isambard-AI National AI Research Resource (AIRR). Isambard-AI is operated by the University of Bristol and is funded by the UK Government’s Department for Science, Innovation and Technology (DSIT) via UK Research and Innovation; and the Science and Technology Facilities Council [ST/AIRR/I-A-I/1023].

\bibliographystyle{IEEEtran}
\bibliography{main}

\begin{IEEEbiography}[{\includegraphics[width=1in,height=1.25in,clip,keepaspectratio]{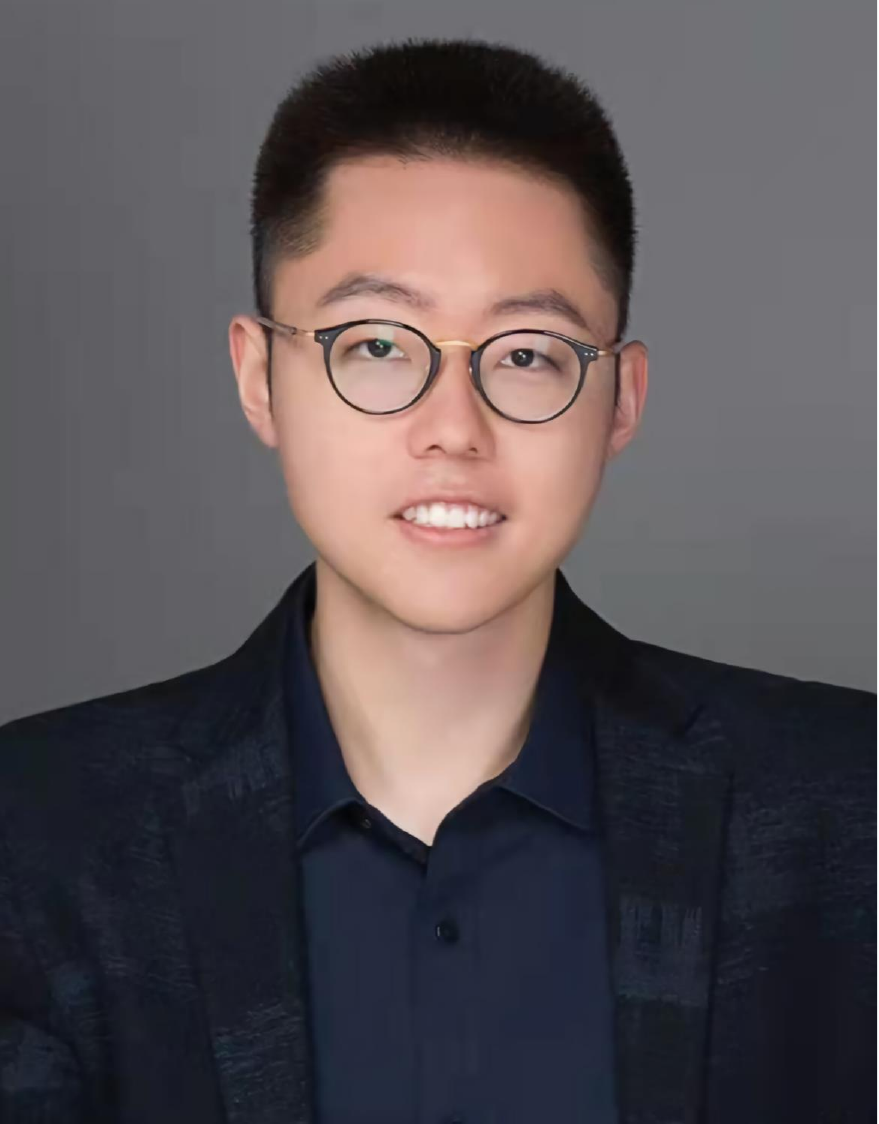}}]{Junpeng Jing} is currently a Research Associate with the Department of Computing, Imperial College London. He received the Ph.D. degree from Imperial College London, U.K., in 2026. He received the  M.Eng. and B.Eng. degrees from Beihang University, China, in 2023 and 2020. His research interests include stereo depth estimation, 3D learning and understanding. In 2023, he won the champion of the Robust Vision Challenge. He has published several papers in top-tier journals and conferences, including IEEE TPAMI, CVPR, ICCV, ECCV, NeurIPS, and ICML, with several recognized as ESI highly cited and highlight papers.
\end{IEEEbiography}
\vfill

\begin{IEEEbiography}[{\includegraphics[width=1in,height=1.25in,clip,keepaspectratio]{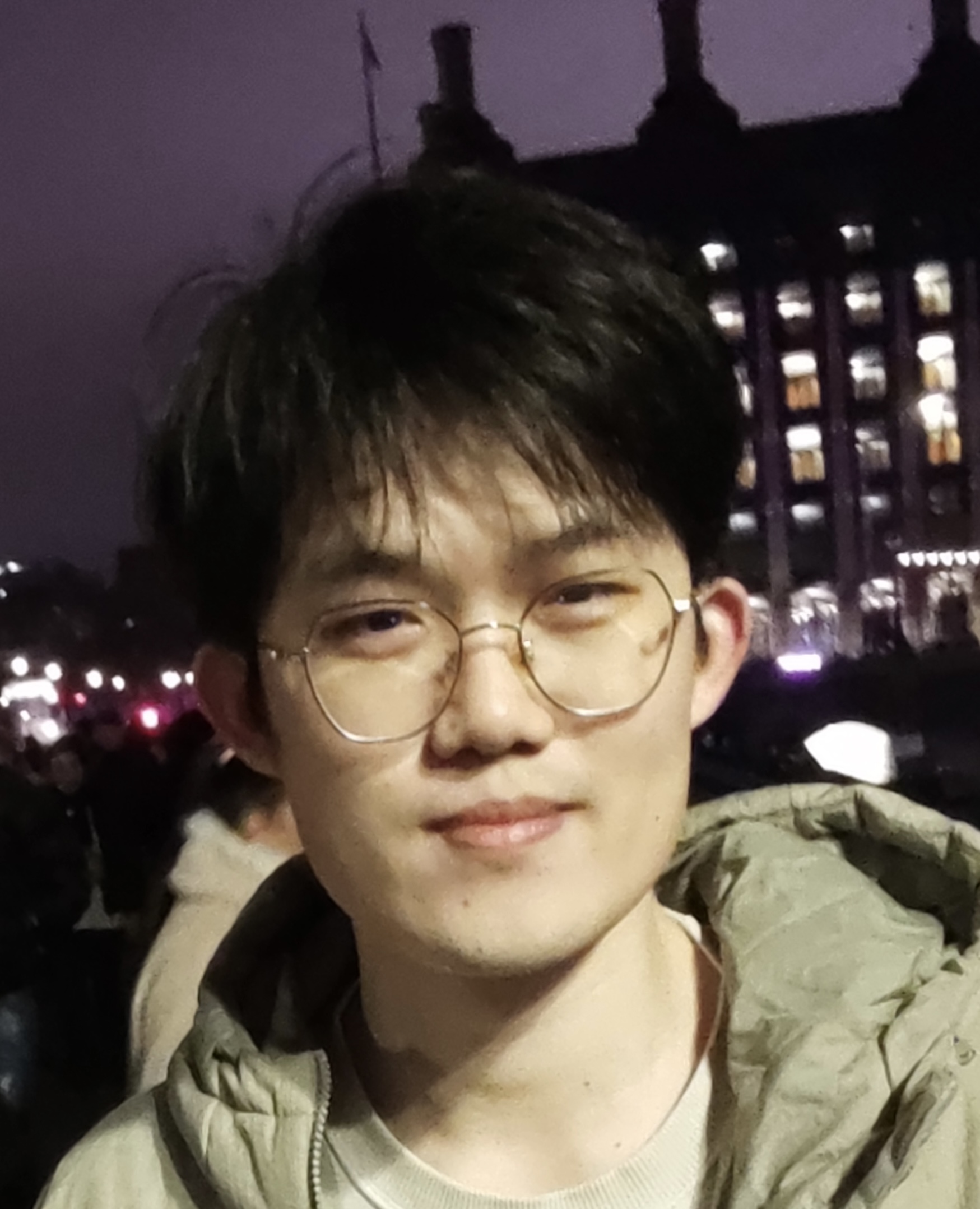}}]{Ronglai Zuo} is currently a Research Associate at Imperial College London. He received his Ph.D. degree from the Hong Kong University of Science and Technology in 2024 and his B.Eng. degree from the Special Class for the Gifted Young, University of Science and Technology of China, in 2020. His research focuses on sign language processing, generative models, and multimodal learning. He has published several papers in top-tier conferences, including CVPR, ICCV, ECCV, and NeurIPS.
\end{IEEEbiography}
\vfill

\begin{IEEEbiography}[{\includegraphics[width=1in,height=1.25in,clip,keepaspectratio]{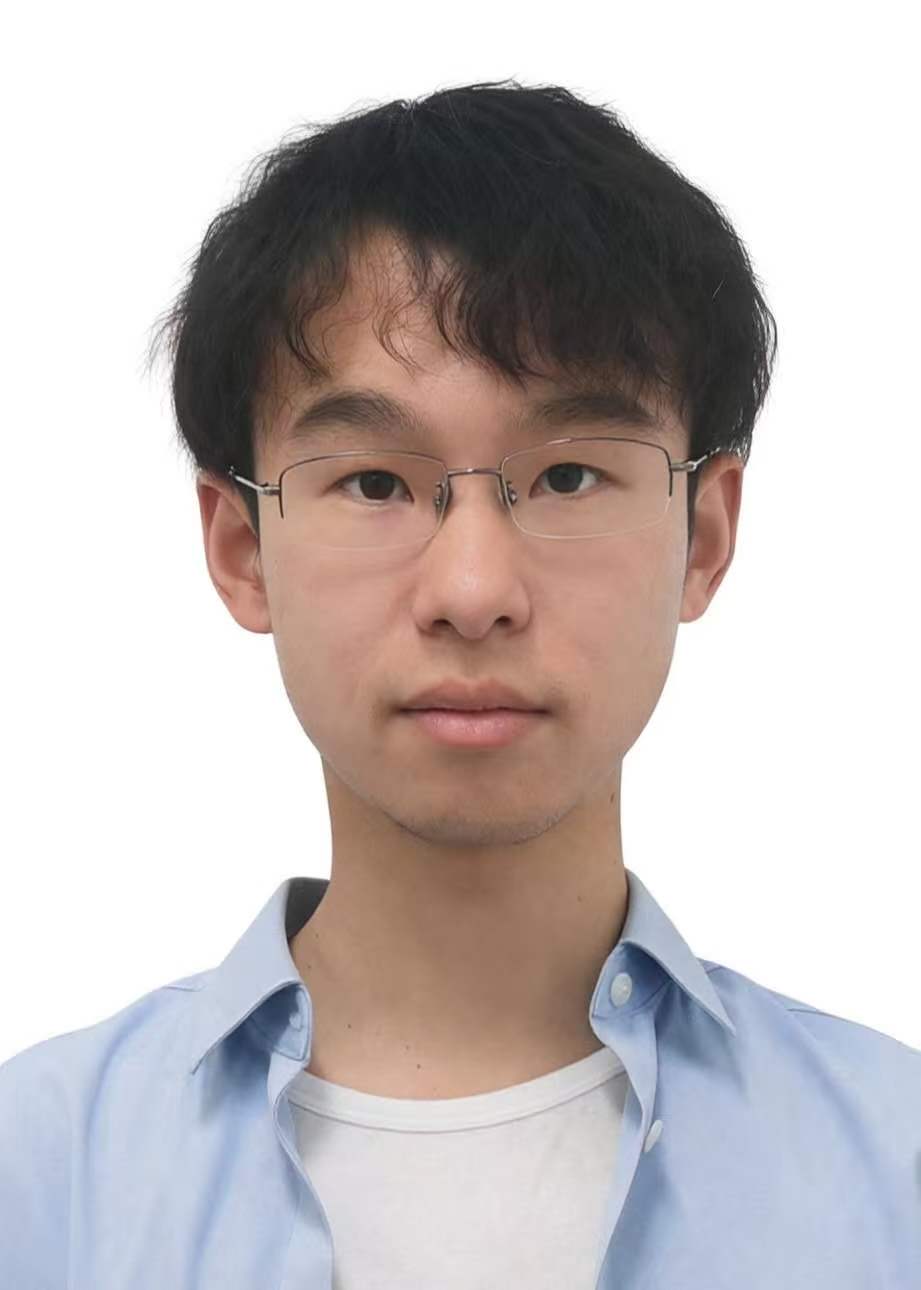}}]{Zhelun Shen} is currently a PhD student at Imperial College London. Prior to this, he was a Senior Researcher at Baidu from 2022 to 2025, where he worked on autonomous driving and generative AI. His research has been published in leading conferences and journals, including CVPR, ECCV, IEEE TPAMI, IEEE TIP, and Pattern Recognition. He won first place in the stereo matching task of the ECCV Robust Vision Challenge (RVC), as well as first place in the Argoverse Stereo Competition at the CVPR 2021 Workshop on Autonomous Driving.
\end{IEEEbiography}
\vfill

\begin{IEEEbiography}[{\includegraphics[width=1in,height=1.25in,clip,keepaspectratio]{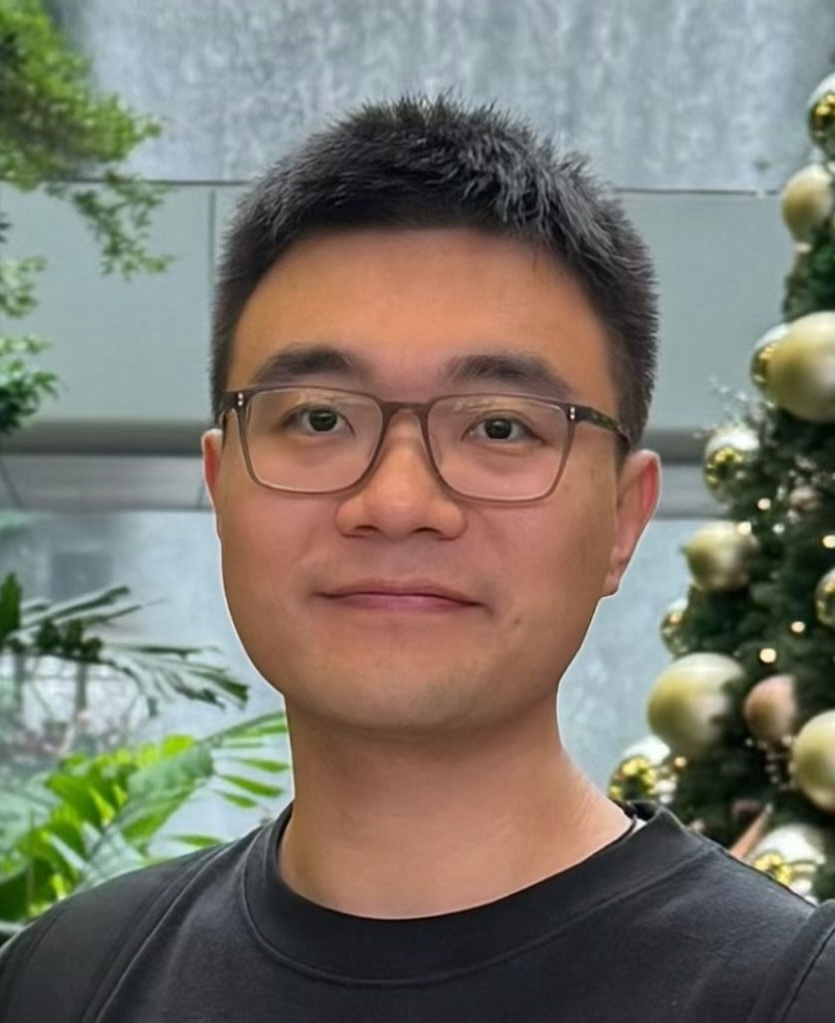}}]{Shangchen Zhou} is currently a Research Associate at Imperial College London. Prior to this, he was a Research Assistant Professor at MMLab@NTU, Nanyang Technological University (NTU), Singapore. He received his Ph.D. (2024) in Computer Science from NTU. He received the NTU CCDS Outstanding PhD Thesis Award in 2025. He won first place in three image restoration and enhancement challenges at NTIRE 2021. His works received notable recognition including the WAIC Youth Outstanding Paper Award Honorable Mention in 2023, the Snap Fellowship Honorable Mention in 2022, and the Best Paper Award at ICIMCS 2016. He also co-organized the MIPI workshop series in conjunction with ECCV 2022, CVPR 2023–2024, and ICCV 2025. He has served as an Area Chair for CVPR, ICLR, and NeurIPS. His research interests include image/video enhancement, generation and editing, etc.
\end{IEEEbiography}
\vfill

\begin{IEEEbiography}[{\includegraphics[width=1in,height=1.25in,clip,keepaspectratio]{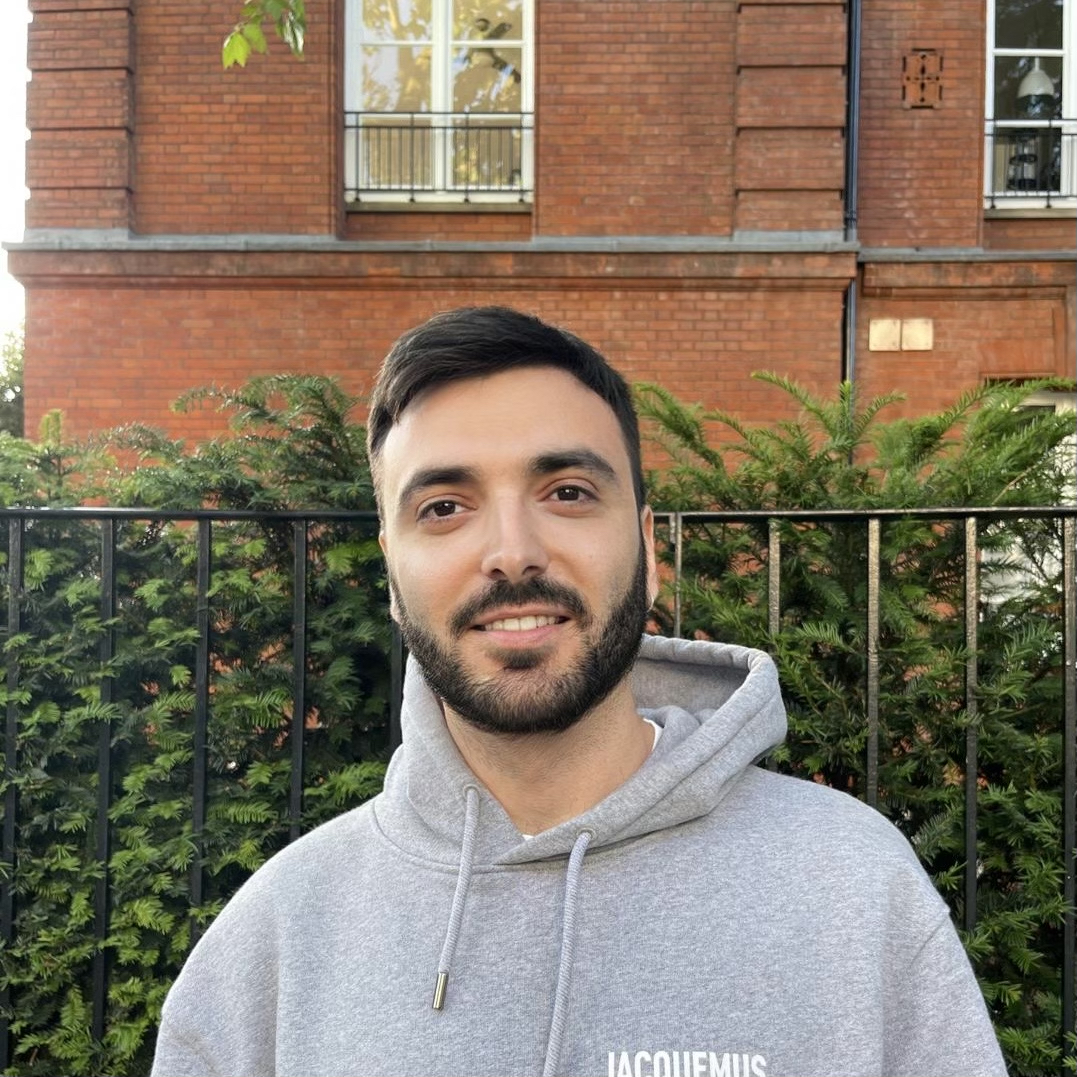}}]{Rolandos Alexandros Potamias} is an Assistant Professor at Imperial College London. He received his PhD from Imperial College London and his M.Eng. degree from National Technical University of Athens. His research is focused on 3D Computer Vision and Embodied AI with a particular focus on human and robot dexterity. He has published several papers in top-tier journals and conferences, including CVPR, ICCV, ECCV, NeurIPS, and ICML.  
\end{IEEEbiography}
\vfill

\begin{IEEEbiography}[{\includegraphics[width=1in,height=1.25in,clip,keepaspectratio]{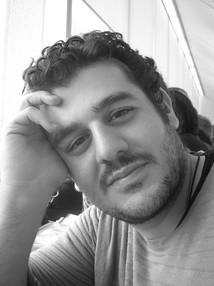}}]{Stefanos Zafeiriou} (Member, IEEE) is a Professor in machine learning and computer vision with the Department of Computing, Imperial College London and a holder of the prestigious UKRI Turing AI World-Leading Research Fellowship. He has co-authored over 250 papers in top-tier machine learning and computer vision venues, including IEEE T-PAMI and IJCV, as well as at leading conferences such as CVPR, ICCV, ECCV, NeurIPS, and ICML. His research focuses on machine learning models applied to computer vision and biosignal analysis. His work has garnered more than 48,000 citations, resulting in an h-index of 92. In recognition of his work, he has received Imperial College’s President’s Medal for Excellence in Research Supervision (2016) and the President’s Medal for Excellence in Innovation and Entrepreneurship (2022). His students are frequent recipients of highly competitive fellowships, such as the Google Fellowship (x2), the Intel Fellowship, and the Qualcomm Fellowship (x4). He has served as an (Guest) Associate Editor for premier journals such as IEEE Transactions on Pattern Analysis and Machine Intelligence (T-PAMI), International Journal of Computer Vision (IJCV), and IEEE Transactions on Affective Computing. He has guest-edited more than eight journal special issues and co-organized over 25 workshops and challenges at top conferences. He also served as the General Chair for BMVC 2017.
\end{IEEEbiography}
\vfill

\begin{IEEEbiography}[{\includegraphics[width=1in,height=1.25in,clip,keepaspectratio]{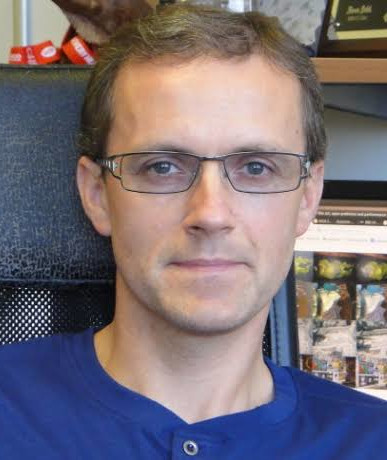}}]{Krystian Mikolajczyk} received the PhD degree from the Institute National Polytechnique de Grenoble and held a number of research positions at INRIA, University of Oxford and Technical University of Darmstadt, as well as faculty positions at the University of Surrey, and Imperial College London. He is a professor at Imperial College London. His main area of expertise is in image and video recognition, in particular methods for image representation and learning. He has served in various roles at major international conferences co-chairing British Machine Vision Conference 2012 and IEEE International Conference on Advanced Video and Signal-Based Surveillance 2013. In 2014 he received Longuet-Higgins Prize awarded by the Technical Committee on Pattern Analysis and Machine Intelligence of the IEEE Computer Society.
\end{IEEEbiography}
\vfill

\begin{IEEEbiography}[{\includegraphics[width=1in,height=1.25in,clip,keepaspectratio]{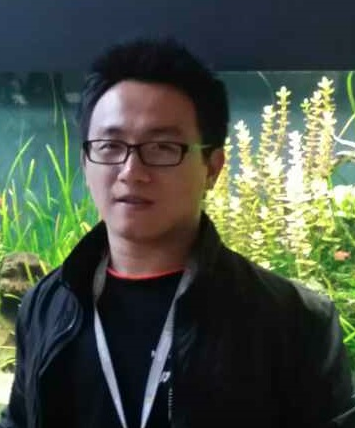}}]{Jiankang Deng} (Member, IEEE) is currently an Assistant Professor with the Department of Computing, Imperial College London. 
His research explores multimodal foundation models and generative modelling of the physical world. He is one of the main contributors to the widely used open-source platform Insightface. He has over 24K citations for his research with an h-index of 52. He is an active area chair of prestigious computer vision and machine learning conferences (e.g., CVPR, ICCV, ECCV, ICML, NeurIPS, ICLR and AAAI). He is also an Associate Editor of IEEE Transactions on Image Processing, Transactions on Machine Learning Research, and Neural Networks. He is a Member of the IEEE.
\end{IEEEbiography}
\vfill

\end{document}